\documentclass[conference]{IEEEtran}
\IEEEoverridecommandlockouts
\usepackage{cite}
\usepackage{amsmath,amssymb,amsfonts}
\usepackage{algorithmic}
\usepackage{graphicx}
\usepackage{textcomp}
\usepackage{xcolor}
\usepackage{tikz}
\usepackage{pgfplots}
\def\BibTeX{{\rm B\kern-.05em{\sc i\kern-.025em b}\kern-.08em
    T\kern-.1667em\lower.7ex\hbox{E}\kern-.125emX}}
\begin{document}

\title{Technical Report on Visual Quality Assessment for Frame Interpolation
}

\author{\IEEEauthorblockN{Hui Men\textsuperscript{1}, Hanhe Lin\textsuperscript{1}, Vlad Hosu\textsuperscript{1}, Daniel Maurer\textsuperscript{2}, Andr{\'e}s Bruhn\textsuperscript{2}, Dietmar Saupe\textsuperscript{1}}\\
\IEEEauthorblockA{\textsuperscript{1}Department of Computer and Information Science, University of Konstanz, Germany\\ \textsuperscript{2}Institute for Visualization and Interactive Systems, University of Stuttgart, Germany}
Email: \{hui.3.men, hanhe.lin, dietmar.saupe\}@uni-konstanz.de, \{Andres.Bruhn, maurerdl\}@vis.uni-stuttgart.de}

\maketitle

\begin{abstract}
Current benchmarks for optical flow algorithms evaluate the estimation quality by comparing their predicted flow field with the ground truth, and additionally may compare interpolated frames, based on these predictions, with the correct frames from the actual image sequences.
For the latter comparisons, objective measures such as mean square errors are applied. However, for applications like image interpolation, the expected user's quality of experience cannot be fully deduced from such simple quality measures. Therefore, we conducted a subjective quality assessment study by crowdsourcing for the interpolated images provided in one of the optical flow benchmarks, the Middlebury benchmark. We used paired comparisons with forced choice and reconstructed absolute quality scale values according to Thurstone's model using the classical least squares method. The results give rise to a re-ranking of 141 participating algorithms w.r.t.\ visual quality of interpolated frames mostly based on optical flow estimation. Our re-ranking result shows the necessity of visual quality assessment as another evaluation metric for optical flow and frame interpolation benchmarks.
\end{abstract}


\section{Introduction}
As one of the basic video processing techniques, frame interpolation, namely computing interpolated in-between images, is a necessary step in numerous applications such as temporal up-sampling for generating slow motion videos, frame rate conversion between broadcast standards and so on \cite{meyer2015phase}. One of the main approaches in frame interpolation is motion compensation, which can be achieved in various ways based on block matching, frequency domain, optical flow, etc.\cite{dufaux1995motion} Among them, optical flow is the most popular one for this usage \cite{meyer2015phase}. Typically, optical flow or its variations are estimated and then used to produce interpolation results. Thus, for a given frame interpolation technique, the quality of the results heavily depends on the optical flow algorithm adopted \cite{niklaus2018context}. 

Regarding the evaluation of motion-compensated interpolation, currently, there is only one optical flow benchmark that is commonly used and offers an assessment of interpolated frames, which is the Middlebury benchmark \cite{baker2011database}. It considers angular and endpoint errors between an estimated flow vector and the ground-truth flow to assess the accuracy of optical flow computation methods. 

\begin{figure}[!t]
\vspace{-10pt}
\centering{\includegraphics[width=0.5\textwidth]{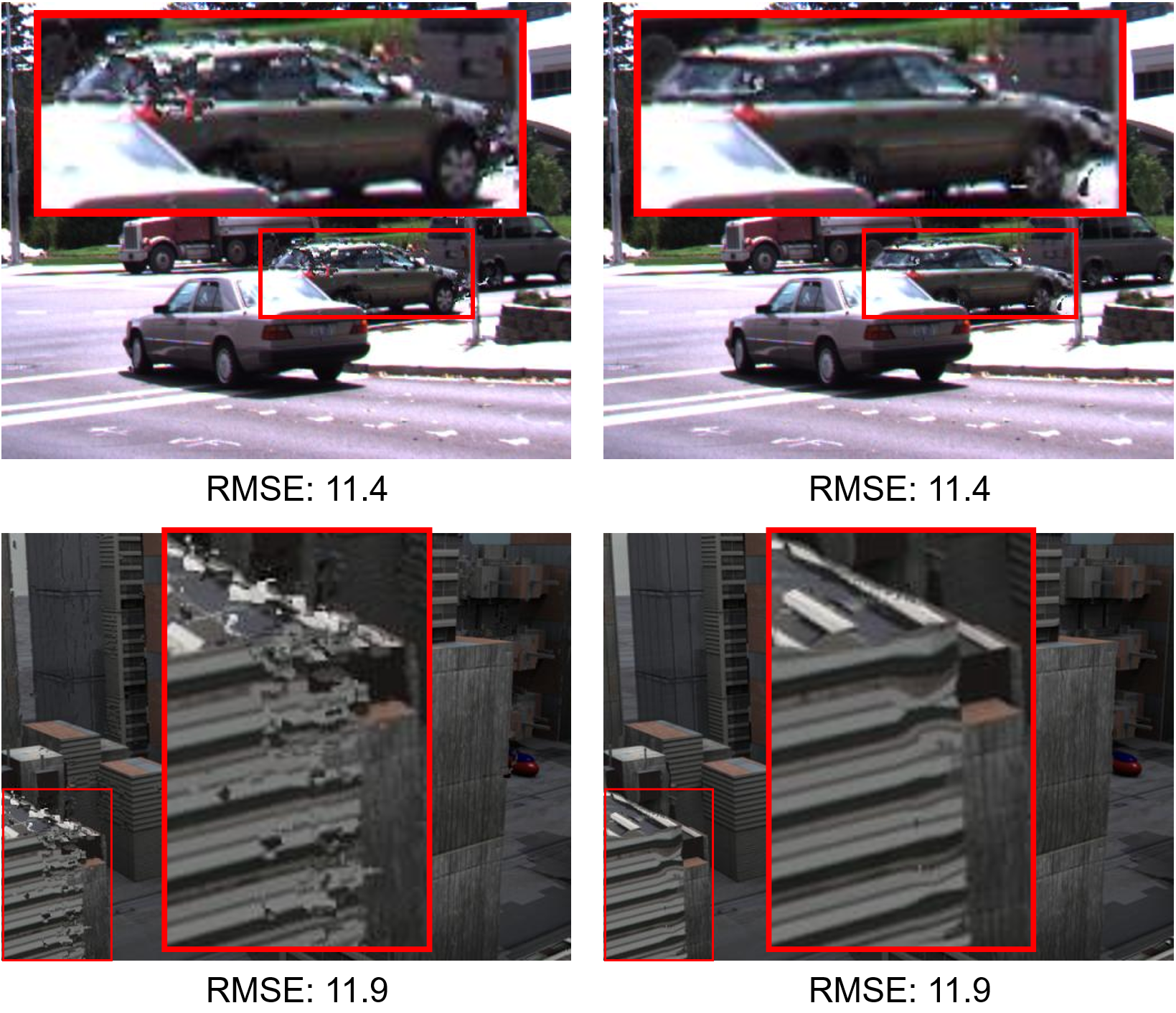}}
\caption{Two interpolated frames, each with two different methods. RMSE values in each pair are equal, but the visual quality differs in each pair, in particular in the zoomed regions.
}
\label{MSE}
%
%
\end{figure}

Moreover, it uses the estimated flow field to interpolate an in-between image for two video frames, and then computes the root mean square error (RMSE) and gradient normalized RMSE between the interpolated image and the ground-truth image. However, it is well known that mean square errors can be misleading and may not reliably reflect image quality as perceived by the human visual system (HVS) \cite{wang2004image}. In the Middlebury interpolation evaluation web-page, some interpolated images have the same RMSE, but exhibit obvious differences in image quality (see Fig.\ref{MSE}). Therefore, we propose that the evaluation of motion-compensated interpolation should take perceived visual quality assessment into consideration. 

\begin{table*}[h]
\caption{SROCC between FR-IQA and Groud-truth (by Subjective Study)}
\label{tb:crd-iqa}
\centering
\begin{tabular}{l |c|c c c c c c c c c c }
  \hline
FR-IQA& Average & Mequon & Schefflera & Urban & Teddy & Backyard & Basketball & Dumptruck & Evergreen\\ 
\hline
RMSE &  0.598	&	0.766	&	0.557	&	0.854	&	0.667	&	0.152	&	0.534	&	0.756	&	0.494 \\
SSIM	& 0.592 	& 0.747	&	0.552	&	0.718	&	0.566	&	0.255	&	0.693	&	0.788	&	0.416	\\
MS-SSIM	&	0.602	&	0.733	&	0.491	&	0.741	&	0.653	&	0.260	&	0.698	&	0.795	&	0.444	\\
FSIM	&	0.599	&	0.739	&	0.573	&	0.783	&	0.631	&	0.244	&	0.553	&	0.778	&	0.488	\\
VSI	&	0.610	&	0.705	&	0.558	&	0.803	&	0.657	&	0.204	&	0.615	&	0.783	&	0.555	\\
  \hline
\end{tabular}
\end{table*}




Regarding visual quality assessment methods, we take full-reference image quality assessment (FR-IQA) into consideration, since ground-truth in-between images are available in the Middlebury benchmark. There are several FR-IQA methods that consider the HVS, such as SSIM \cite{wang2004image}, MS-SSIM \cite{wang2003multi}, FSIM \cite{zhang2011fsim}, VSI \cite{zhang2014vsi}. These methods were designed to estimate image quality degradation due to common artifacts, namely the ones caused by processing such as data compression or by losses in data transmission. However, the artifacts induced by optical flow algorithms lead to interpolated images with different specific distortions. 

In this contribution, we show (in Table \ref{tb:crd-iqa}) that five objective FR-IQA methods have rather low correlations with the evaluations made by human observers, regardless of whether the methods are based on the HVS or just on pixel-wise errors such as RMSE. More specifically, the method VSI, one of the best FR-IQA methods, when trained and tested on the LIVE database \cite{sheikh2006statistical}, yielded a Spearman rank-order correlation coefficient (SROCC) of 0.952 w.r.t.\ ground truth consisting of MOS from a controlled lab study for the LIVE database. But when applied for the interpolated images by optical flow algorithms, VSI gave only an SROCC of 0.610. This shows that current FR-IQA methods do not perform well on interpolated frames produced by optical flow algorithms because of their specific distortions. Therefore, a new FR-IQA method specifically designed for such images is needed. However, before the research in such FR-IQA methods can proceed, ground-truth data, namely subjective quality scores of such images need to be collected, a first set of which is provided by our study.

For subjective quality evaluation, lab-studies take the lead because of their reliability. In lab-studies, the experimental environment and the evaluation process can be controlled. However, it is time consuming and costly, which severely limits the number of images to be assessed. Crowdsourcing studies can be less expensive and the reliability of crowdsourcing has been proven to be acceptable with appropriate setup and certain training for the crowd workers \cite{saupe2016crowd}. Therefore, we applied subjective quality assessment of the images interpolated by optical flow algorithms with the help of crowdsourcing. 

In this paper, we implemented paired comparisons of the interpolated images given by optical flow algorithms in the Middlebury interpolation benchmark with the help of crowdsouring and re-ranked them accordingly. After that, the old ranking according to RMSE in the Middlebury benchmark and the re-ranking according to our subjective study were compared. After that, the old ranking according to RMSE in the Middlebury benchmark and the re-ranking according to our subjective study were compared. Our study shows that
current FR-IQA methods are not suitable for assessing the perceived visual quality of interpolated frames (as produced by optical flow algorithms).

\section{Related Work}
So far, there is only one benchmark that is used for evaluating the performance of frame interpolation, namely the Middlebury benchmark. It was originally designed for the evaluation of optical flow algorithms. Since it provides the ground-truth in-between images to evaluate the interpolation performance of optical flow algorithms, some interpolation algorithms  also made use of this benchmark for evaluation, \cite{raket2012motion}, \cite{meyer2015phase}, \cite{niklaus2018context}.  

Some interpolation algorithms like \cite{liu2017video}, \cite{gongvideo} used the UCF 101 dataset \cite{soomro2012ucf101} for training and testing. Others like \cite{raket2012motion},  \cite{zhai2005low}, \cite{ghutke2016novel} used the videos from \cite{xidian1}, \cite{xidian2}. For evaluation, generally they chose to compute one of MSE, PSNR, and SSIM between their interpolated images and the ground-truth in-between images.

\section{Subjective Study using Paired Comparisons}

\subsection{Subjective Study}

\emph{Absolute Category Rating} (ACR) is a judgment where the test items are presented one at a time and are rated independently on a category of five ordinal scales, i.e., Bad-1, Poor-2, Fair-3, Good-4, and Excellent-5 \cite{itu1999subjective}. 

ACR is easy and fast for implementation, however, it has several problems \cite{chen2009crowdsourceable}. Participants may be confused when the concept of the ACR scale has not been explained sufficiently well. They may also have different interpretations of the ACR scale, in particular in crowdsourcing experiments because of the wide range of backgrounds and perceptual experiences of the crowd workers from around the world. Moreover, the perceptual distances between two consecutive scale values, e.g., between 1 and 2, should ideally be the same. However, in practice this can hardly be achieved \cite{hossfeld2016qoe}. Also it is not easy to detect when a participant intentionally or carelessly gives false ratings.

Alternatively, {\em paired comparisons} (PC) can solve some of the problems of ACR. In a PC test, items to be evaluated are presented as pairs. In a forced choice setting, one of the items must be chosen as the preferred one. The main advantage of this is that it is highly discriminatory, which is very relevant when test items have nearly the same quality.

However, when implemented naively, to compare $N$ items would require ${N \choose 2 }$ comparisons, too many to be practical, when $N$ is on the order of 100, for example. In our case, for each of the 8 sequences, we would have to compare $N=141$ images, giving a total of 78,960 pairs. 

Instead, random paired comparisons randomly chooses a fraction of all possible paired comparisons. This is more efficient and has been proven to be as reliable as full comparisons \cite{xu2012hodgerank}. After obtaining paired comparisons results, subjective quality scores can be reconstructed based on Thurstone's model \cite{thurstone1927law}, \cite{luce1994thurstone} or the Bradley-Terry model \cite{bradley1952rank}. 

\subsection{Thurstone's Model} 

Thurstone's model provides the basis for a psychometric method for assigning scale values to options on a one-dimensional continuum from paired comparisons data. It assumes that an option's quality is a Gaussian random variable, thereby accommodating differing opinions about the quality of an option.
Then each option's latent quality score is revealed by the mean of the corresponding Gaussian. 

The result of a paired comparison experiment is a square count matrix $C$ denoting the number of times that each option was preferred over any other option. More specifically, for $n$ comparisons of option $A_i$ with option $A_j$, $C_{i,j}$ gives the number of times $A_i$ was preferred over $A_j$. Similarly, $C_{j,i}$ in the count matrix denotes the number of times that $A_j$ was preferred over $A_i$, and we have $C_{i,j}+ C_{j,i}=n$. 

According to Thurstone' Case V, subjective qualities about two options A and B are modelled as uncorrelated Gaussian random variables $A$ and $B$ with mean opinions $\mu_A,\mu_B$ and variances ${\sigma_A}^2,{\sigma_B}^2$, respectively. When individuals decide which of the two options is better, they draw realizations from their quality distributions, and then choose the option with higher quality. More specifically, they choose option A over option B if their draw from the random variable $A-B$ (with mean $\mu_{AB} = \mu_{A}  - \mu_{B}$ and variance ${\sigma_{AB}}^2 = {\sigma_{A}}^2 + {\sigma_{B}}^2$) is greater than $0$. Therefore, the probability of a subject to prefer option A over B is: \begin{equation} \label{eq1} P(A > B) = P(A − B > 0)= \Phi \left( \frac{ \mu_{AB} }{ \sigma_{AB} } \right),\end{equation} where $\Phi(\cdot)$ is the standard normal cumulative distribution function (CDF). 

Thurstone proposed to estimate $P(A > B)$ by the empirical proportion of people preferring A over B, which can be derived from the count matrix $C$ as 
$$
    P(A>B) \approx \frac{C_{A,B}}{C_{A,B}+C_{B,A}}.
$$
The estimated quality difference $\hat{\mu}_{AB}$ can be derived from inverting Eq.\ \ref{eq1}, giving: $$\hat{\mu}_{AB}=\sigma_{AB} \Phi ^ {-1} \left( \frac{C_{A,B}}{C_{A,B}+C_{B,A}} \right)$$ 
known as Thurstone’s Law of Comparative Judgment,  where $\Phi(\cdot)^{-1}$ is the inverse standard normal CDF, or z-score. Least-squares fitting or maximum likelihood estimation (MLE) can be then applied to estimate the scale values $\mu_{A}$ for all involved stimuli $A$. For more details we refer to the technical report \cite{tsukida2011analyze}.

\subsection{Study Design}

\begin{figure}[t!]
\centering{\includegraphics[width=0.5\textwidth]{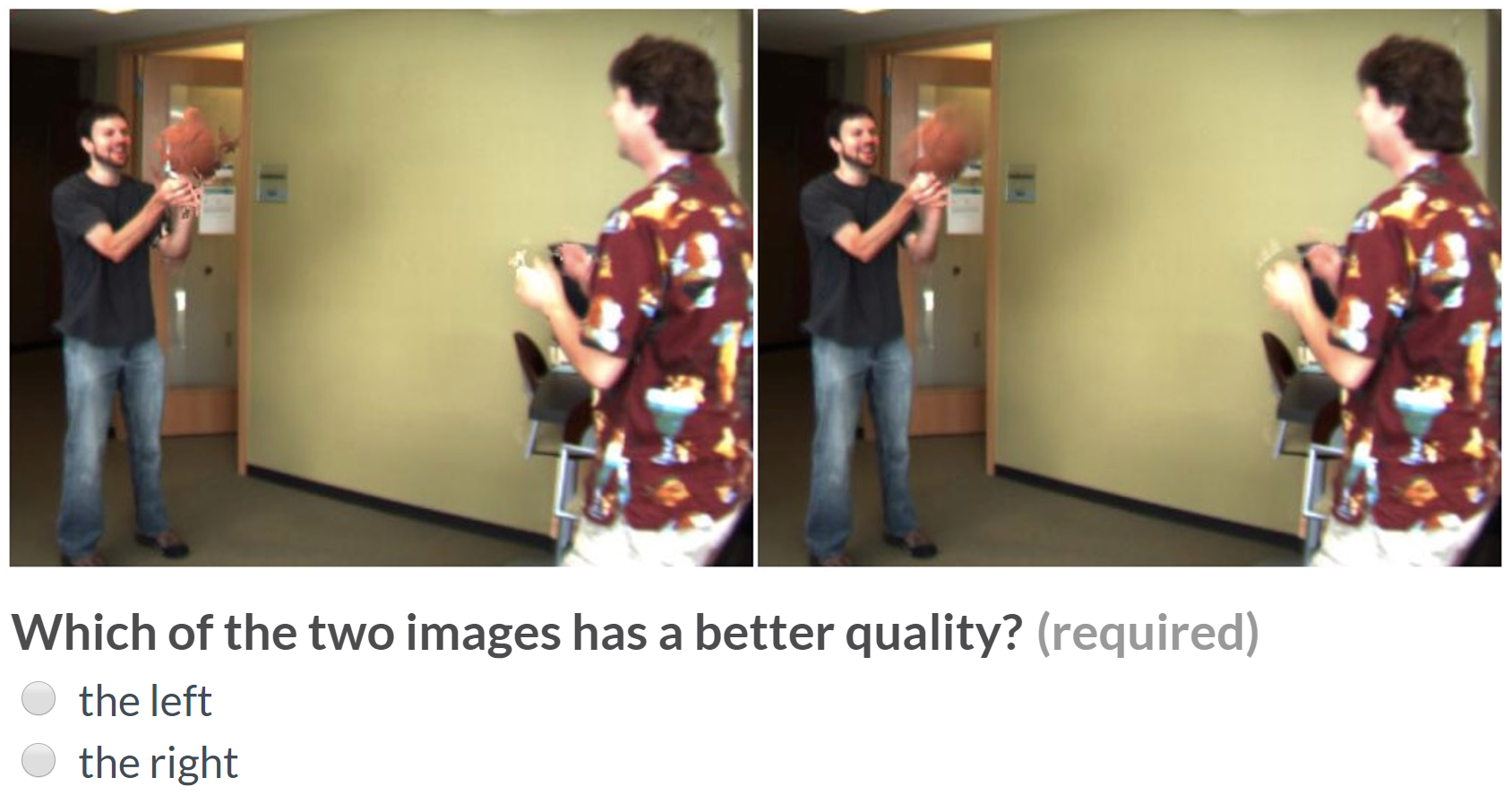}}
\caption{Crowdsourcing interface.}
\label{crwins}
\end{figure}

In order to re-rank the methods in the Middlebury benchmark, we implemented paired comparisons based on Thurstone's model with least-squares estimation to obtain subjective judgments of the image qualities. In the benchmark, there are 8 sets of interpolated images, most of which generated by 141 optical flow methods.\footnote{Note that when we ran the experiments in June 2018, there were altogether 141 methods in the Middlebury benchmark, which now includes a number of additional, new methods.} Therefore, in our experiment, for each set of 141 interpolated images, we generated a random sparse graph with degree of 6 (i.e., each image was to be randomly compared to 6 other images), which resulted in 423 pairs of images. We ran the experiment using the Figure Eight \cite{plat} platform. In our crowdsourcing interface as shown in Fig.\ \ref{crwins}, crowd workers were asked to identify and select the image with better quality for each image pair (forced binary choice).

\begin{figure}[t!]
\centering{\includegraphics[width=0.5\textwidth]{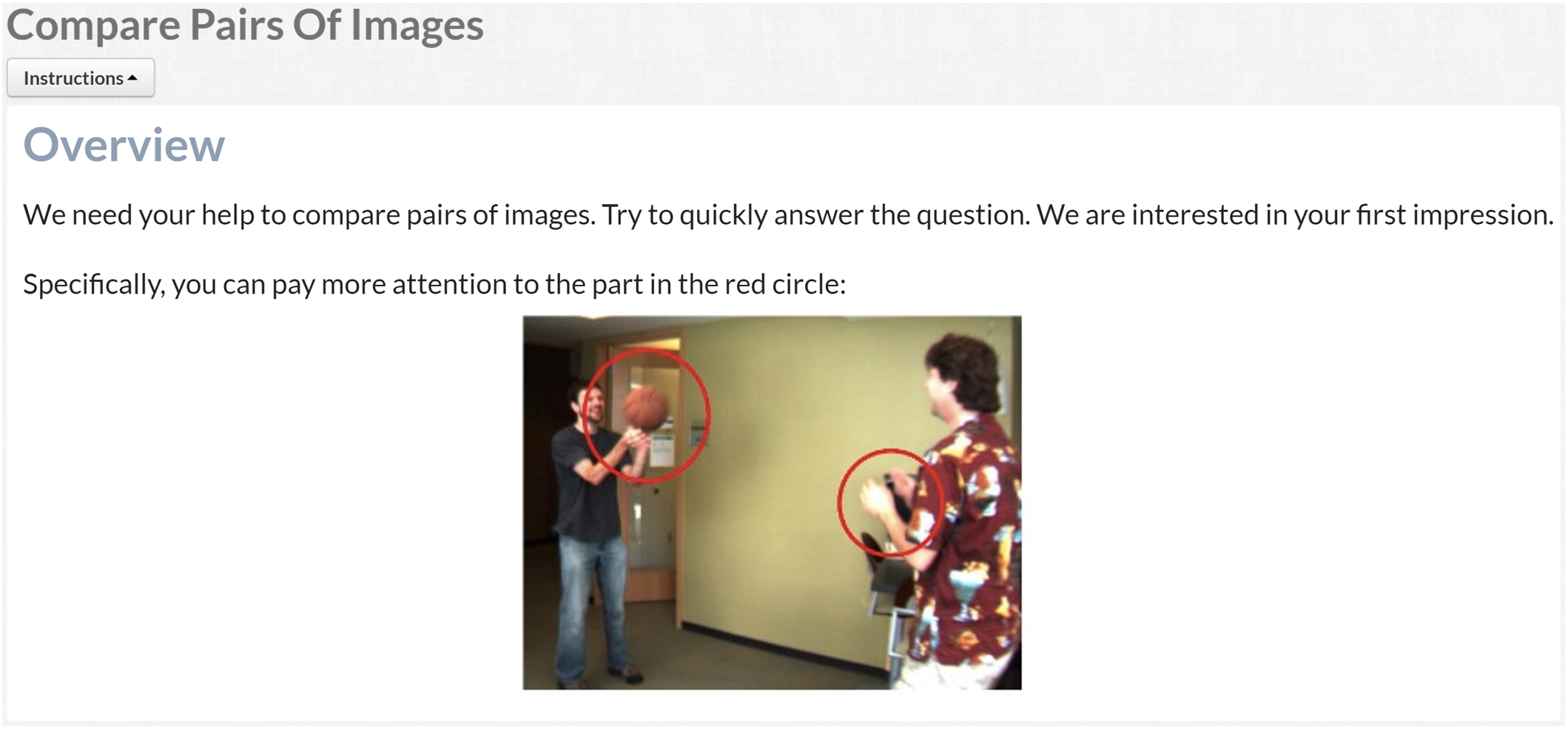}}
\caption{Instructions of the crowdsourcing experiment.}
\label{crdins2}
\end{figure}

\subsection{Quality Assurance and Quality Control}
Before the actual paired comparison, there was a training session, in which workers were instructed how to compare the image quality. Since the visual differences between some images are not that obvious, in the instructions as shown in Fig.\ \ref{crdins2}, we highlighted those parts in the images that are more degraded and hence showed more differences.

In our experiment, we had eight separate jobs, each containing the 423 image pairs of the same series. In each job, there were 22 tasks, each consists of one page. In order to make sure that crowd workers' performance was not effected by exhaustion, we showed 20 pairs of images per page, and payments were initiated for each completed page. For each pair of images to be compared, we collected 30 votes from the crowd workers. 

In order to assure the reliability of the crowd workers, the unreliable ones need to be detected and disallowed to continue. This was done by requiring crowd workers to answer test questions. For the test questions we chose image pairs with the ground-truth in-between image as one of the images, and the other image of bad quality. Then the expected, correct answer was obviously given by choosing the ground-truth image as the one with a better quality. 

Before crowd workers were allowed to start a job, they must pass a quiz which is composed entirely of test questions. This ensures that only crowd workers that proved to be able to work on the subject matter of the job, would be able to enter the job. Crowd workers that failed the quiz were permanently disqualified from working on the job. After passing the quiz, crowd workers were admitted to start the real job. During the job, they had to answer further hidden test questions. Once a crowd worker failed more than 30\% of the hidden test questions, he or she was disqualified and removed from the job. Only crowd workers who passed the quiz and showed an accuracy above 70\% on the hidden test questions were regarded as reliable.

\section{Results}
\subsection{Statistics for the Crowdsourcing Experiment}
In our experiment, there were eight jobs, each one comparing 141 interpolated images pairwise (423 pairs per job). The average run time was 29 hours per job. In total, 3189 crowd workers participated in our experiment, some of them took multiple jobs. Before the real experiment, 54\% of them did not pass the quiz thus were not allowed to contribute to the job. During the real job, 14\% of them failed more than 30\% of the test questions, thus been disregarded. In the end, 1033 crowd workers were accepted as trusted workers. Among the trusted workers, 79\% had an accuracy of 90\%-100\% (where 100\% means they passed all test questions), 10\% of them had an accuracy of 80\%-90\% and 10\% of them had an accuracy of 70\%-80\%.

\begin{table*}[h]
\caption{Bootstrapped Correlations between RMSE and Ground-truth (by Subjective Study) after 1000 Iterations}
\label{tb:corrbootstrp}
\centering
\resizebox{1\textwidth}{!}{
\begin{tabular}{l |c|c c c c c c c c c c }
  \hline
RMSE& Average & Mequon & Schefflera & Urban & Teddy & Backyard & Basketball & Dumptruck & Evergreen\\ 
\hline
SROCC	& 0.598		&	0.766	&	0.557	&	0.854	&	0.667	&	0.152	&	0.534	&	0.756	&	0.494	\\
CI 95\%	& 	[0.507,0.674]	&	[0.699,0.816]	&	[0.454,0.647]	&	[0.813,0.888]	&	[0.581,0.737]	&	[0.015,0.283]	&	[0.419,0.618]	&	[0.695,0.813]	&	[0.382,0.593]	\\
	\hline																	
\end{tabular}
}
\end{table*}

\subsection{Re-ranking Results}
Given the results of the paired comparisons, we reconstructed corresponding quality values based on Thurstone's model using the code provided by \cite{tsukida2011analyze}. In order to make the results of the 8 separate jobs comparable, we added two fictitious images as anchors. One of them represents the worst quality among all the images, and the other one is like the ground-truth image, with a quality that is better than that of all the other images. After reconstruction of the scale values for the 141 + 2 images in each series, we linearly rescaled the quality scores such that the quality of the imaginary worst quality image became 0, and that of the ground-truth image became 1. In this way, we rescaled the reconstructed scores to the interval $[0,1]$. Reconstructed quality values are shown in Table \ref{tb:value1} and \ref{tb:value2}, accompanied by their corresponding rankings.
Note that for each set, we ranked them separately. The `Average' in the first column was gained by taking the mean of the 8 quality values in the same row, which results in the rank according to average quality.

Table \ref{tb:rankcom1} and \ref{tb:rankcom2} show the differences between the re-ranking (ranked according to subjective study) and their corresponding ranking in the Middlebury benchmark (ranked according to RMSE), denoted by `new' and `old' in the tables, respectively. It can be seen that the best three methods ranked by the subjective study (i.e., SuperSlomo, CtxSyn and DeepFlow2), ranked 1st, 5th and 9th in the Middlebury benchmark, respectively. Overall, 36 methods highlighted in color blue showed rank differences up to 5. However, 30 methods highlighted in color red gave differences of more than 30 between their new and old rankings.


\begin{figure}[h]
\centering{\includegraphics[width=0.5\textwidth]{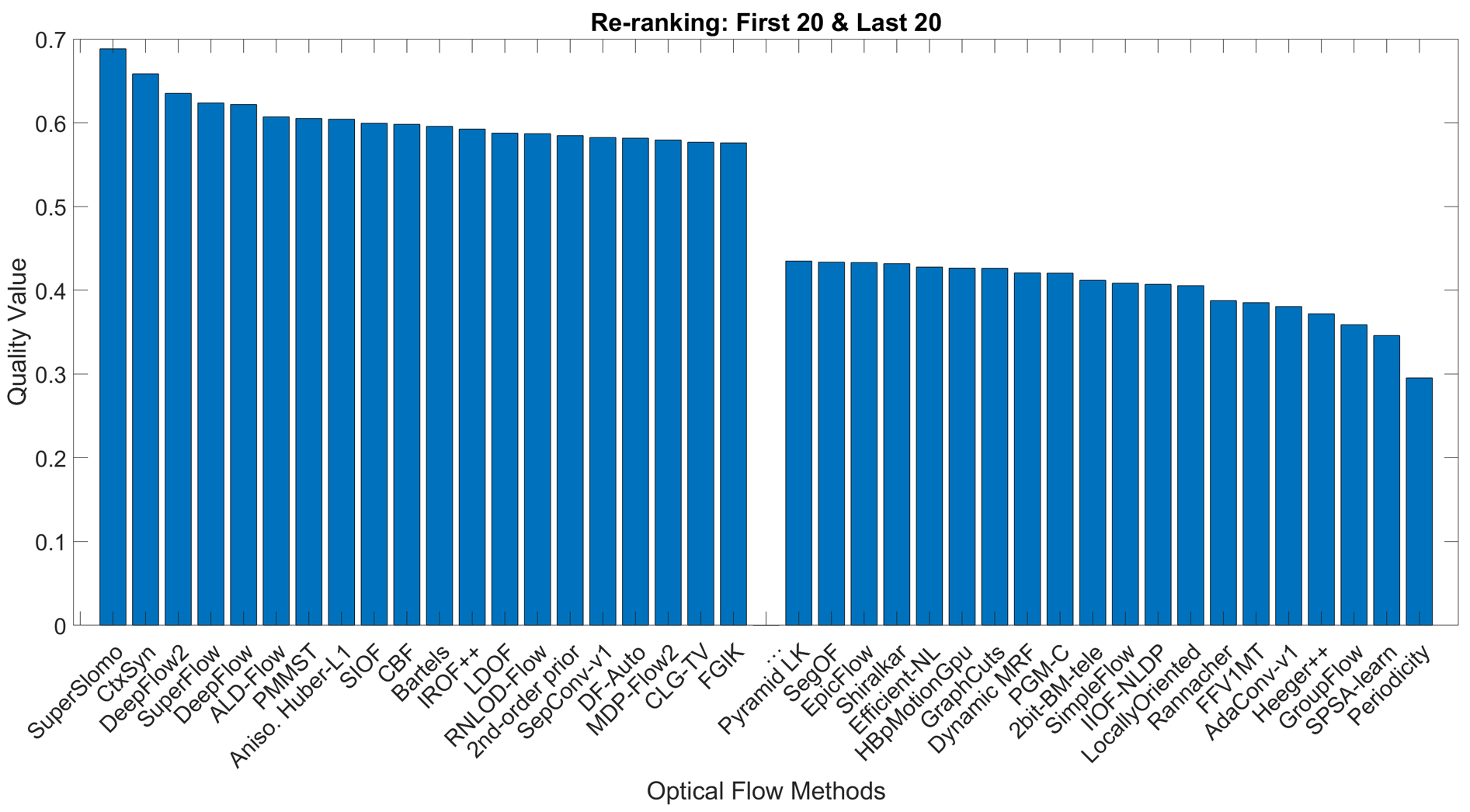} }
\caption{The methods ranking in the top 20 and the bottom 20 by the subjective study. The x-axis shows the names of the methods. The y-axis denotes the value of the average subjective scores.}
\label{bar}
\end{figure} 


\begin{figure*}[h!]
\centering{\includegraphics[width=0.95\textwidth]{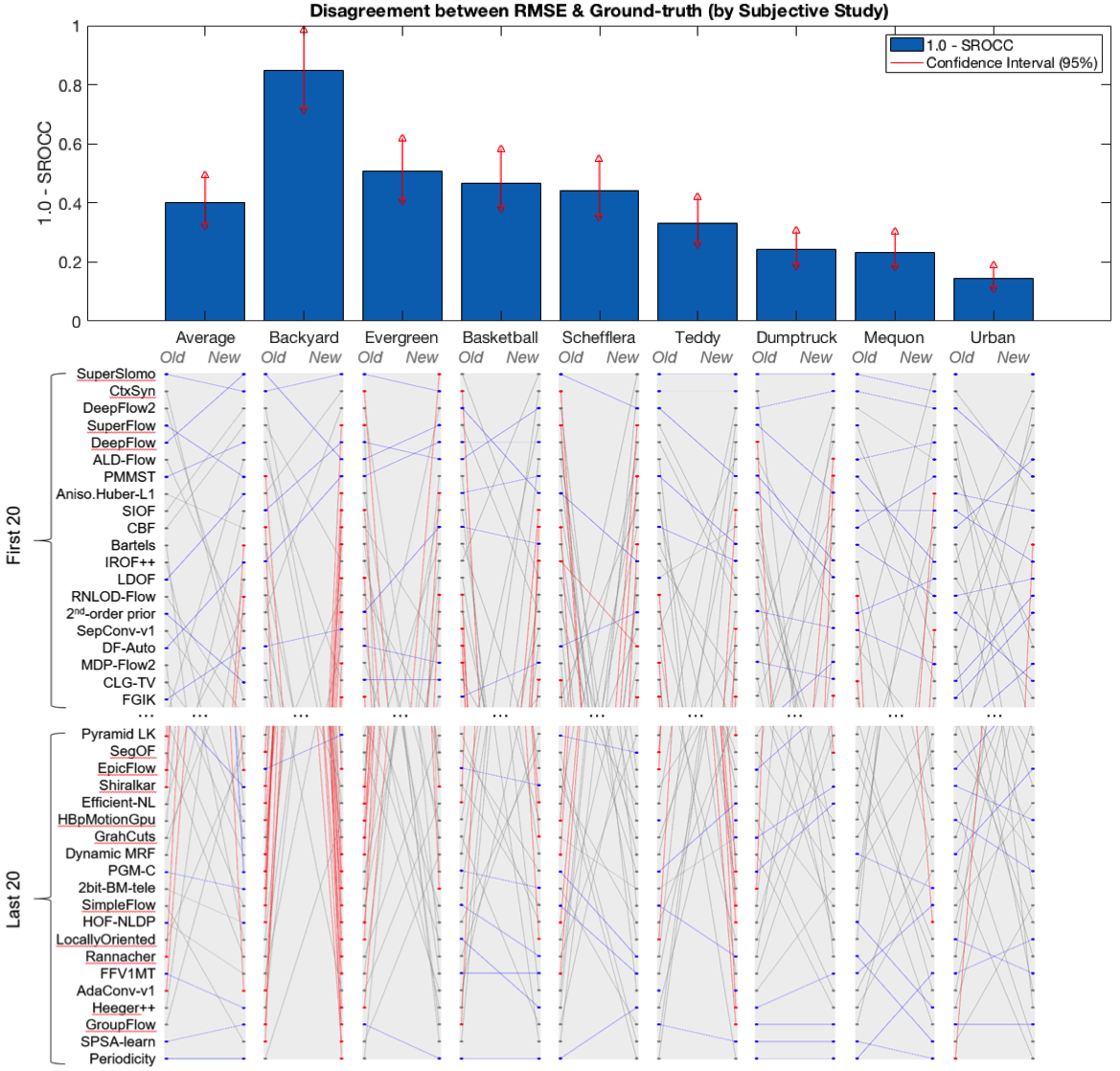}}
\caption{Disagreement level and corresponding ranking differences. Upper bars: disagreement level with 95\% confidence interval. Lower graphs: nodes on the left side denote their old ranking in the Middlebury benchmark Nodes on the right side denote their new ranking by subjective study. Lines in blue color denote the ranking differences are less then 5, and lines in color red denote the ranking differences are larger than 50.}
\label{visual}
\end{figure*}

\begin{figure*}[h!]
\centering{\includegraphics[width=0.25\textwidth]{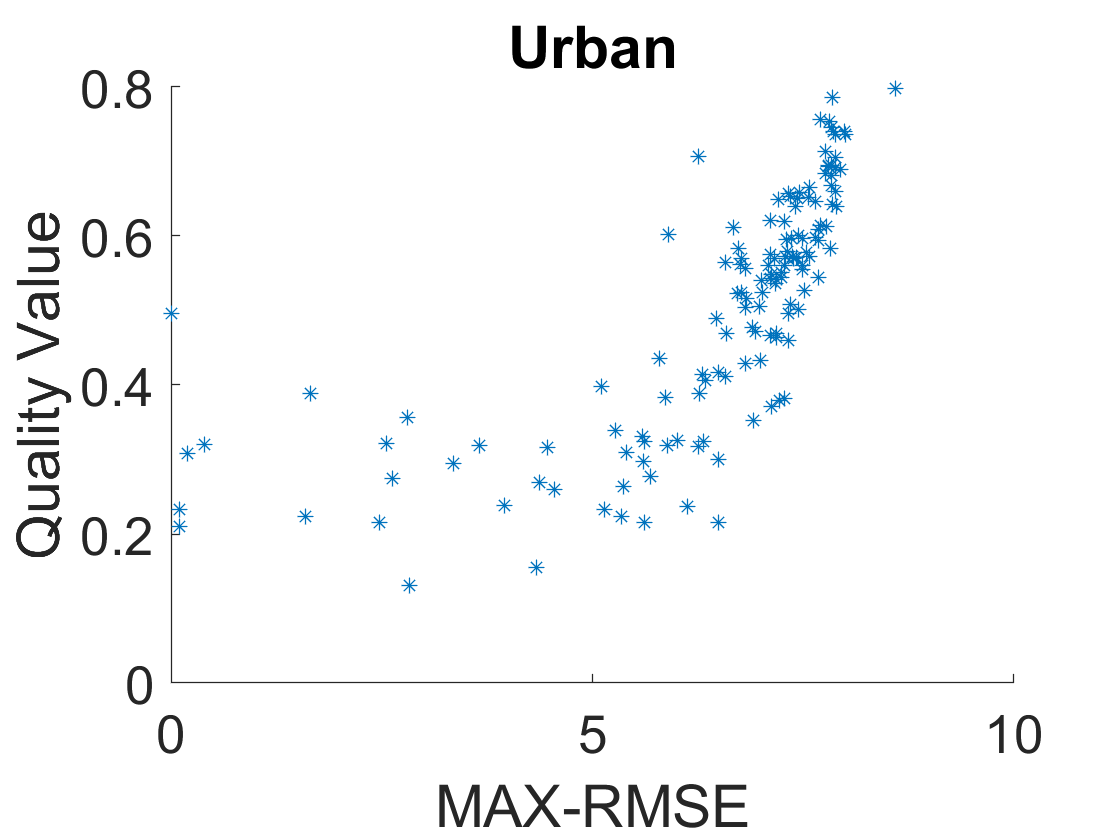}\includegraphics[width=0.25\textwidth]{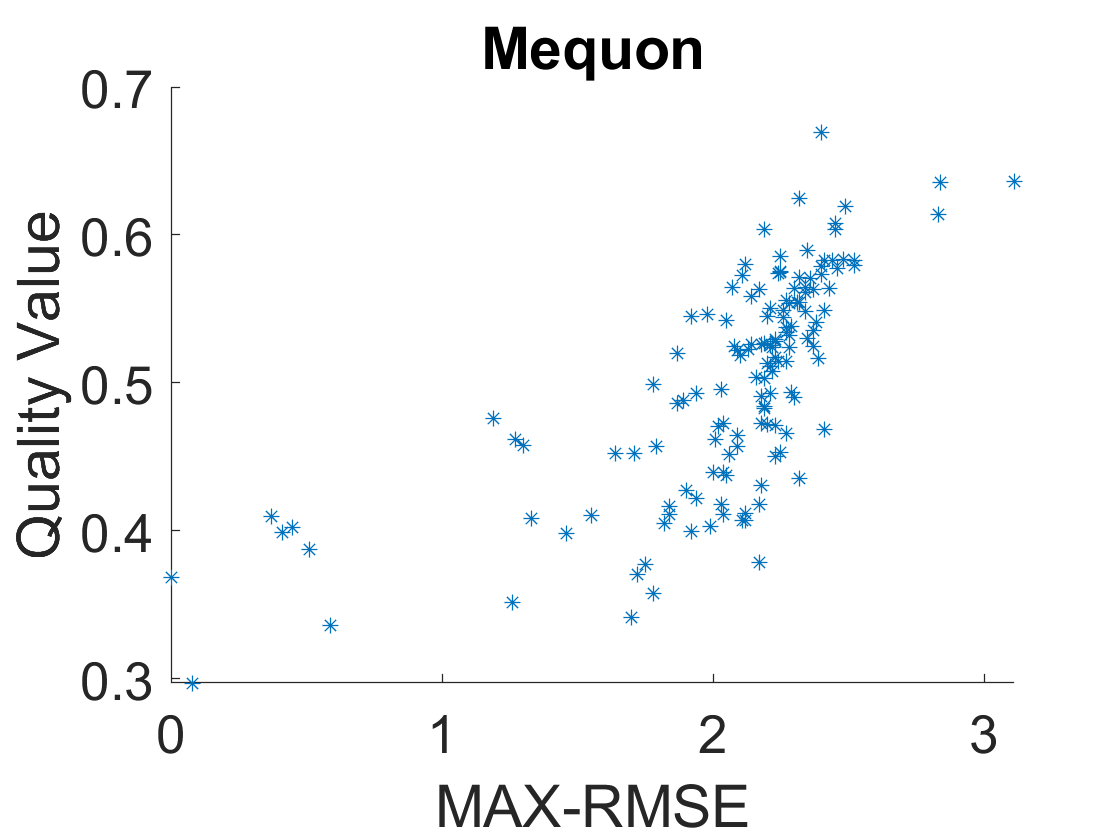}\includegraphics[width=0.25\textwidth]{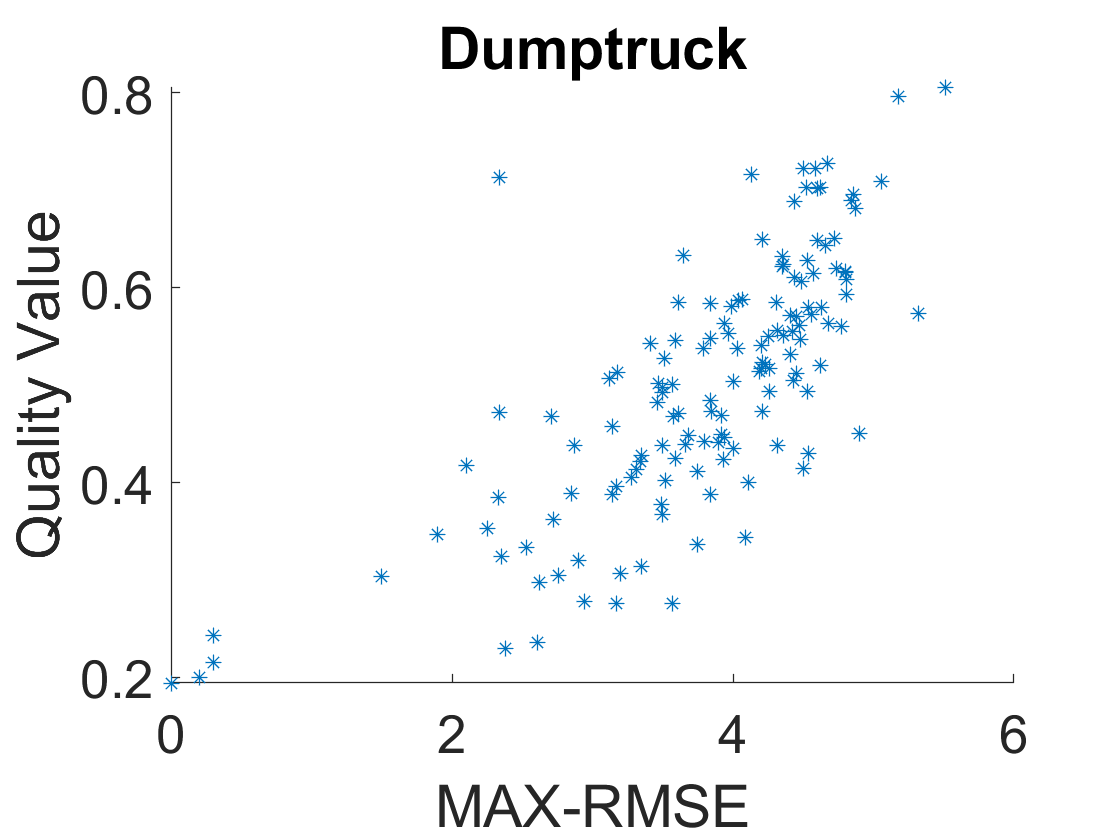}\includegraphics[width=0.25\textwidth]{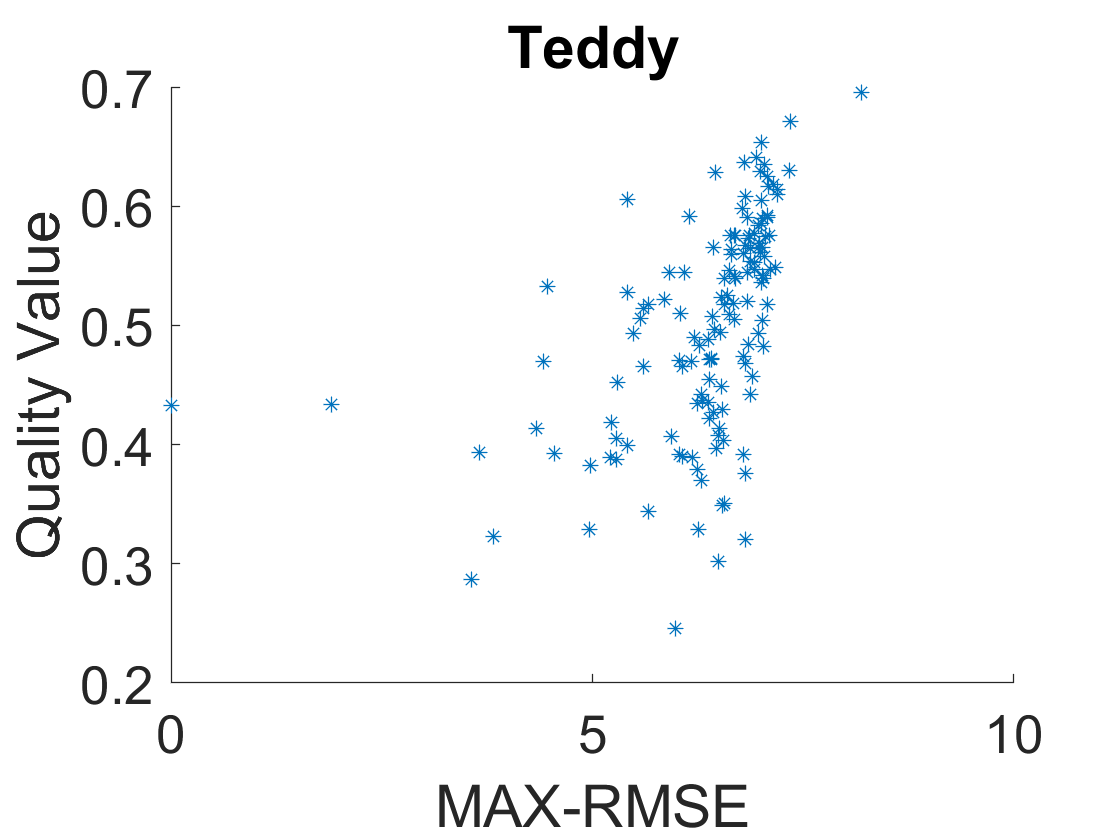} \\ \vspace{6pt} \includegraphics[width=0.25\textwidth]{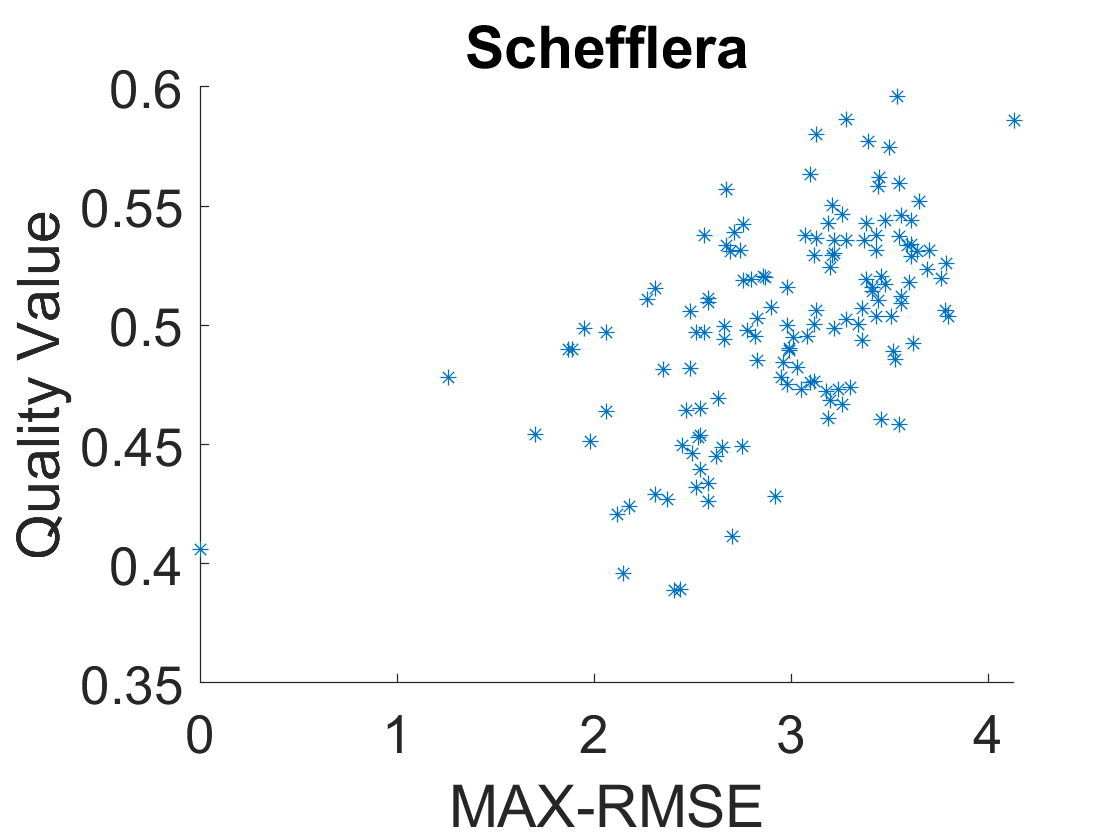}\includegraphics[width=0.25\textwidth]{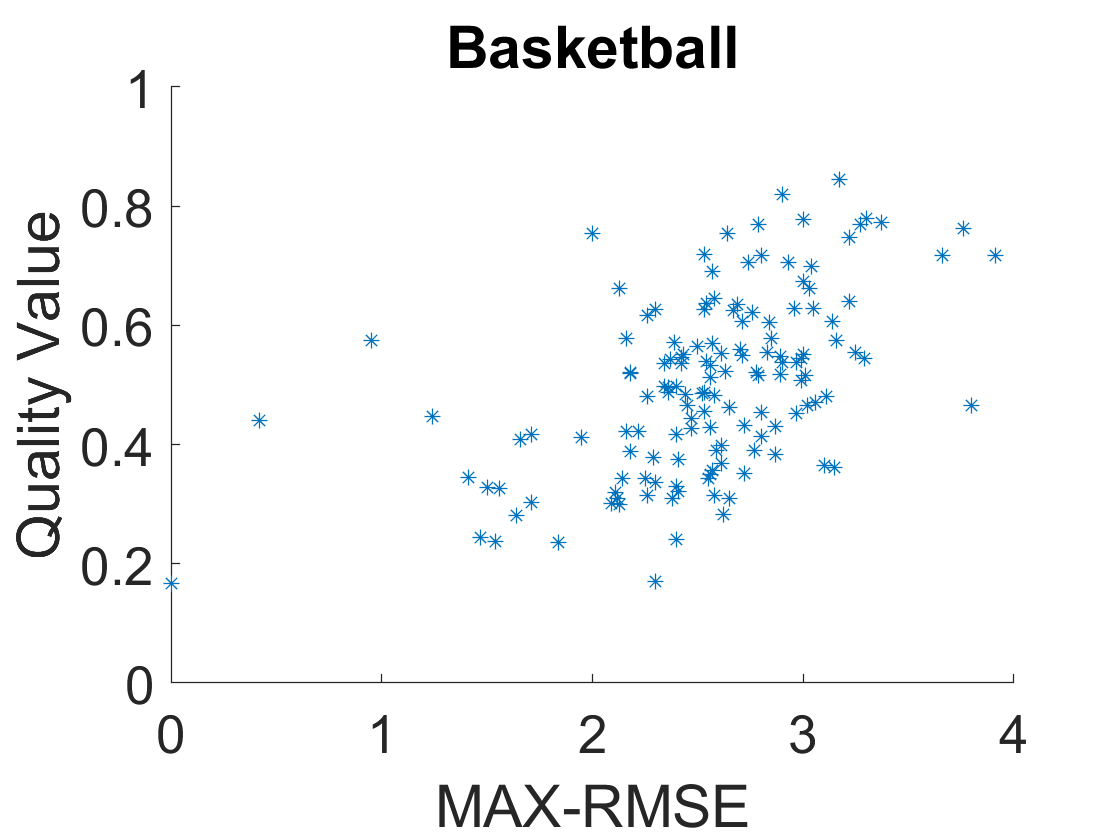}\includegraphics[width=0.25\textwidth]{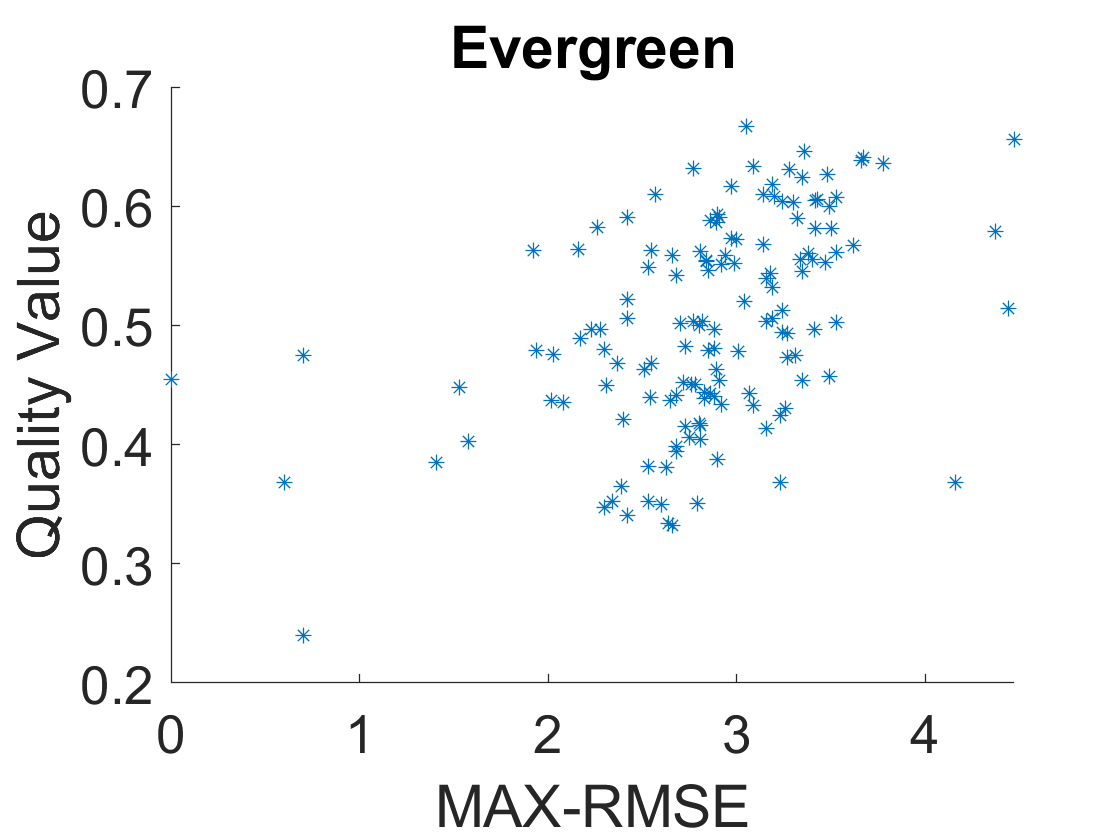}\includegraphics[width=0.25\textwidth]{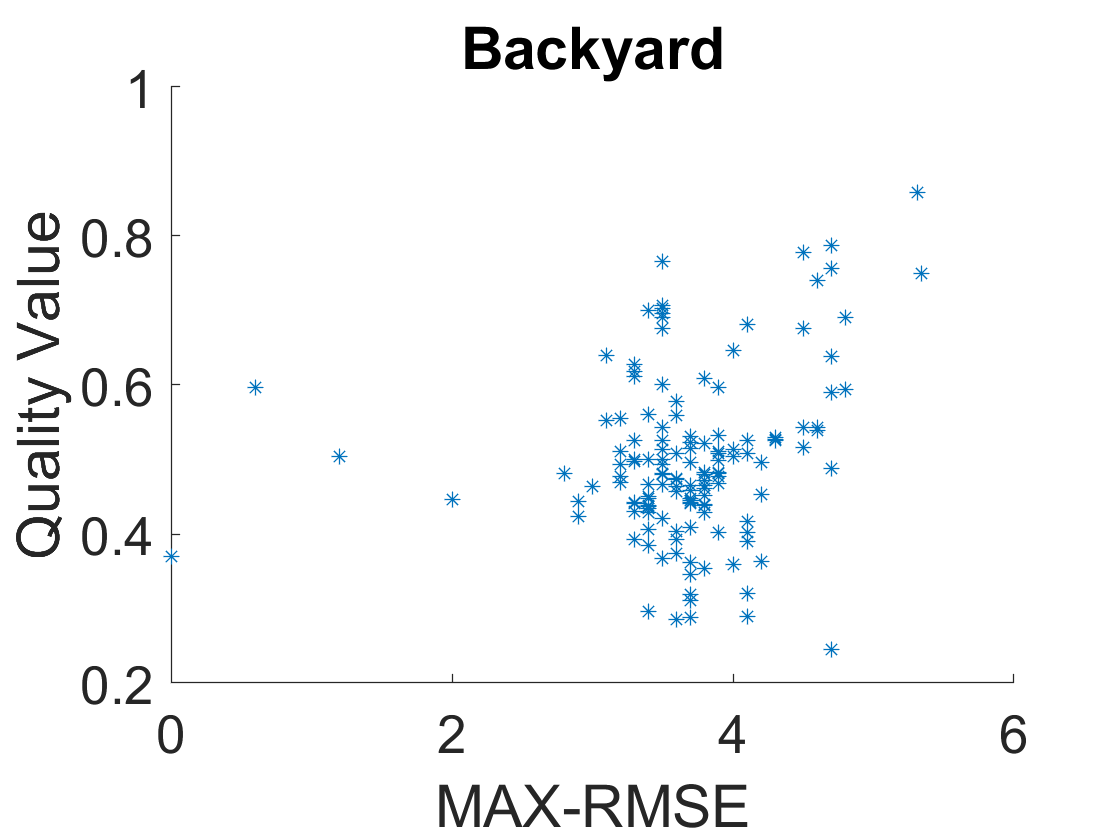}}
\caption{Scatter plots of RMSE and subjective scores. To show positive correlations, we used the difference between maximum and individual RMSE as the x-axis values.} 
\label{scatter plot}
\end{figure*} 


Fig.\ \ref{bar} shows the subjective qualities of the methods ranked the highest 20 and the lowest 20. The quality scores of the best two methods, SuperSlomo and CtxSyn, are better than those of the rest by a large margin.  

As an overall analysis, Table \ref{tb:corrbootstrp} shows the bootstrapped (after 1000 iterations) SROCC correlation values accompanied with confidence intervals (95\%) 
between the ranking in Middlebury benchmark (i.e., ranking according to RMSE) and the re-ranking according to our subjective study. Note that the confidence interval of SROCC was computed by transforming the rank correlation score into an approximate z-score using the Fisher transform \cite{ruscio2008constructing}. In a nutshell, a confidence interval of probability $p$ is given by
$\tanh(\arctan r \pm \Phi^{-1}(p) /\sqrt{n-3})$, where $r$ denotes the estimated SROCC, $n$ is the sampling size, and $\Phi^{-1}$ is the inverse of the standard normal CDF. In order to visualize the result, we computed the disagreement level as $1 - SROCC$ as shown in Fig.\ \ref{visual}. 

Fig.\ \ref{scatter plot} shows the scatter plots of the RMSE values compared to subjective scores of each optical flow method. Those eight plots are displayed in descending order of SROCC between RMSE and subjective scores. It can be seen that starting from \emph{Urban} with the highest SROCC (0.854) down to the lowest SROCC (0.152) given by \emph{Backyard}, the scattered plots become more sparse, as to be expected from their decreasing correlation.




\begin{table*}[h!]
\caption{Resolutions of the Original Images and the Available Ones used for Subjective Study}
\label{tb:resolution}
\centering
\begin{tabular}{c |c c c c c c c c c c }
  \hline
  & Mequon & Schefflera & Urban & Teddy & Backyard & Basketball & Dumptruck & Evergreen\\ 
\hline
Original &  584$\times$388 & 584$\times$388  & 640$\times$480 & 420$\times$360 & 640$\times$480 & 640$\times$480 & 640$\times$480 & 640$\times$480 \\
Available     	&	467$\times$310	&	467$\times$310	&	512$\times$384	&	336$\times$288	&	512$\times$384	&	512$\times$384	&	512$\times$384	&	512$\times$384	\\
  \hline
\end{tabular}
\end{table*}


\section{Limitations}
The interpolated images available from the Middlebury benchmark are compressed and slightly downsized from the original images. The original interpolated images could not be made available by the maintainers of the Middlebury benchmark. The differences between their resolutions are shown in Table~\ref{tb:resolution}. 
Since we used the downscaled, compressed public version of the images for the crowdsourcing study, our results may be biased to a small extent. 

Another limitation of the experiment is the difficulty of the subjective study. The quality differences between some images are quite hard to distinguish. Therefore, in the instructions of the crowdsourcing experiment, we highlighted the main degraded parts according to our visual observation to help the crowd workers to focus on the critical parts of the images. We believe, that this can be further improved in future studies, e.g., by providing zoomed image portions that contain the most noticeable artifacts.


\section{Conclusion and Future Work} 
We have adopted visual quality assessment to the Middlebury benchmark for frame interpolation based mostly on optical flow methods. Our study confirms that only using RMSE as an evaluation metric for image interpolation performance is not representative of visual quality. Also current FR-IQA methods do not provide satisfying results on those interpolated images. This is due to the fact that such images, especially the ones generated by optical flow algorithms have specific distortions that are quite different from  artifacts commonly addressed by conventional IQA methods. 

Therefore, we plan to develop a domain specific FR-IQA for frame interpolation based on optical flow estimation. For the reference there is the original ground truth frame. In addition, we can make use of the ground truth optical flow vector field which is also available for the frame to be interpolated. This amounts to a FR-IQA with side information given by optical flow. It requires feature extraction from images and additionally from the optical flow, in order to train a model for a FR-IQA method specifically for frame interpolation by means of optical flow. The use case of such a FR-IQA method is to serve as a visual quality metric in optical flow benchmarks. Moreover, we plan to apply VQA methods on the videos generated by frame interpolation as a further study. This, in turn, will allow us to consider temporal aspects in the quality assessment.

\section*{Acknowledgment}
Funded by the Deutsche Forschungsgemeinschaft (DFG, German Research Foundation) -- Projektnummer 251654672~-- TRR 161 (Project A05 and B04).

\bibliographystyle{IEEEtran}
\bibliography{bib}

\newpage

\appendices

\begin{table*}[h]
\caption{Subjective Quality Values and the Re-ranking of the Middlebury Benchmark (Part I). }
\label{tb:value1}
\centering
\begin{tabular}{l |l|l l l l l l l l l  }
  \hline
 & \multicolumn{1}{c}{Average}\vline & \multicolumn{1}{c}{Mequon} & \multicolumn{1}{c}{Schefflera} & \multicolumn{1}{c}{Urban}& \multicolumn{1}{c}{Teddy}& \multicolumn{1}{c}{Backyard}& \multicolumn{1}{c}{Basketball}& \multicolumn{1}{c}{Dumptruck}& \multicolumn{1}{c}{Evergreen}\\
    & value / rank & value / rank &	value / rank	&	value / rank	&	value / rank	&	value / rank	&	value / rank	&	value / rank	&	value / rank\\
        \hline
SuperSlomo 	&	0.688	/	1	&	0.635	/	3	&	0.504	/	71	&	0.798	/	1	&	0.671	/	2	&	0.750	/	6	&	0.763	/	8	&	0.806	/	1	&	0.579	/	32	\\
CtxSyn 	&	0.658	/	2	&	0.636	/	2	&	0.586	/	3	&	0.464	/	91	&	0.695	/	1	&	0.858	/	1	&	0.718	/	13	&	0.796	/	2	&	0.515	/	61	\\
DeepFlow2 	&	0.635	/	3	&	0.571	/	26	&	0.530	/	39	&	0.736	/	9	&	0.590	/	23	&	0.596	/	26	&	0.780	/	3	&	0.648	/	18	&	0.631	/	9	\\
SuperFlow 	&	0.624	/	4	&	0.583	/	13	&	0.476	/	104	&	0.501	/	82	&	0.554	/	46	&	0.756	/	5	&	0.777	/	4	&	0.703	/	10	&	0.638	/	5	\\
DeepFlow 	&	0.622	/	5	&	0.535	/	51	&	0.550	/	13	&	0.692	/	16	&	0.605	/	17	&	0.532	/	38	&	0.747	/	11	&	0.709	/	8	&	0.603	/	21	\\
ALD-Flow 	&	0.607	/	6	&	0.565	/	27	&	0.586	/	2	&	0.660	/	22	&	0.609	/	15	&	0.417	/	116	&	0.845	/	1	&	0.696	/	12	&	0.479	/	82	\\
PMMST 	&	0.605	/	7	&	0.608	/	7	&	0.534	/	31	&	0.690	/	17	&	0.504	/	77	&	0.638	/	19	&	0.772	/	5	&	0.593	/	32	&	0.502	/	68	\\
Aniso. Huber-L1 	&	0.604	/	8	&	0.579	/	18	&	0.534	/	32	&	0.614	/	35	&	0.617	/	12	&	0.521	/	47	&	0.641	/	24	&	0.702	/	11	&	0.626	/	10	\\
SIOF 	&	0.599	/	9	&	0.538	/	49	&	0.531	/	36	&	0.657	/	24	&	0.586	/	25	&	0.690	/	12	&	0.662	/	21	&	0.563	/	45	&	0.568	/	35	\\
CBF 	&	0.598	/	10	&	0.580	/	17	&	0.529	/	40	&	0.705	/	12	&	0.610	/	14	&	0.593	/	27	&	0.575	/	40	&	0.611	/	29	&	0.582	/	30	\\
Bartels 	&	0.596	/	11	&	0.486	/	86	&	0.538	/	22	&	0.683	/	18	&	0.524	/	64	&	0.740	/	7	&	0.495	/	73	&	0.727	/	3	&	0.573	/	33	\\
IROF++ 	&	0.592	/	12	&	0.625	/	4	&	0.575	/	6	&	0.541	/	71	&	0.630	/	7	&	0.477	/	77	&	0.769	/	6	&	0.644	/	19	&	0.478	/	83	\\
LDOF 	&	0.588	/	13	&	0.571	/	25	&	0.482	/	98	&	0.468	/	89	&	0.640	/	4	&	0.787	/	2	&	0.706	/	17	&	0.484	/	80	&	0.561	/	41	\\
RNLOD-Flow 	&	0.587	/	14	&	0.493	/	80	&	0.580	/	4	&	0.621	/	33	&	0.575	/	33	&	0.700	/	10	&	0.547	/	52	&	0.622	/	24	&	0.558	/	44	\\
2nd-order prior 	&	0.585	/	15	&	0.583	/	14	&	0.503	/	72	&	0.609	/	38	&	0.536	/	60	&	0.609	/	23	&	0.508	/	70	&	0.722	/	5	&	0.608	/	16	\\
SepConv-v1 	&	0.582	/	16	&	0.614	/	6	&	0.486	/	95	&	0.496	/	83	&	0.629	/	8	&	0.488	/	68	&	0.717	/	15	&	0.574	/	41	&	0.656	/	2	\\
DF-Auto 	&	0.582	/	17	&	0.469	/	95	&	0.563	/	7	&	0.598	/	41	&	0.542	/	55	&	0.778	/	3	&	0.629	/	28	&	0.494	/	78	&	0.581	/	31	\\
MDP-Flow2 	&	0.579	/	18	&	0.577	/	19	&	0.492	/	89	&	0.668	/	20	&	0.614	/	13	&	0.590	/	28	&	0.551	/	49	&	0.689	/	13	&	0.454	/	95	\\
CLG-TV 	&	0.577	/	19	&	0.549	/	41	&	0.485	/	96	&	0.583	/	47	&	0.635	/	6	&	0.480	/	74	&	0.554	/	46	&	0.703	/	9	&	0.624	/	11	\\
FGIK 	&	0.576	/	20	&	0.583	/	15	&	0.465	/	113	&	0.324	/	112	&	0.606	/	16	&	0.676	/	15	&	0.719	/	12	&	0.713	/	7	&	0.521	/	59	\\
IROF-TV 	&	0.572	/	21	&	0.532	/	53	&	0.519	/	50	&	0.693	/	15	&	0.618	/	11	&	0.509	/	54	&	0.622	/	32	&	0.532	/	61	&	0.553	/	49	\\
LME 	&	0.572	/	22	&	0.574	/	23	&	0.516	/	54	&	0.601	/	40	&	0.589	/	24	&	0.515	/	49	&	0.606	/	36	&	0.682	/	15	&	0.493	/	76	\\
TC/T-Flow 	&	0.572	/	23	&	0.526	/	59	&	0.543	/	19	&	0.754	/	4	&	0.494	/	79	&	0.452	/	94	&	0.820	/	2	&	0.425	/	103	&	0.559	/	43	\\
Modified CLG 	&	0.569	/	24	&	0.583	/	12	&	0.426	/	134	&	0.653	/	25	&	0.590	/	21	&	0.681	/	14	&	0.391	/	104	&	0.620	/	25	&	0.604	/	20	\\
CombBMOF 	&	0.563	/	25	&	0.484	/	87	&	0.560	/	9	&	0.560	/	61	&	0.598	/	18	&	0.496	/	65	&	0.706	/	16	&	0.555	/	50	&	0.545	/	54	\\
CRTflow 	&	0.562	/	26	&	0.549	/	40	&	0.531	/	34	&	0.509	/	79	&	0.518	/	68	&	0.467	/	82	&	0.699	/	18	&	0.633	/	20	&	0.590	/	25	\\
p-harmonic 	&	0.561	/	27	&	0.590	/	10	&	0.519	/	51	&	0.640	/	31	&	0.575	/	32	&	0.510	/	53	&	0.480	/	80	&	0.628	/	22	&	0.544	/	55	\\
TV-L1-MCT 	&	0.559	/	28	&	0.431	/	112	&	0.530	/	38	&	0.524	/	76	&	0.539	/	57	&	0.497	/	63	&	0.629	/	27	&	0.716	/	6	&	0.605	/	19	\\
OAR-Flow 	&	0.558	/	29	&	0.524	/	64	&	0.467	/	112	&	0.786	/	2	&	0.637	/	5	&	0.458	/	91	&	0.516	/	68	&	0.522	/	64	&	0.552	/	50	\\
FMOF 	&	0.557	/	30	&	0.604	/	8	&	0.534	/	30	&	0.352	/	108	&	0.575	/	29	&	0.542	/	35	&	0.754	/	9	&	0.572	/	42	&	0.520	/	60	\\
NNF-Local 	&	0.554	/	31	&	0.564	/	30	&	0.526	/	42	&	0.642	/	30	&	0.576	/	28	&	0.539	/	37	&	0.550	/	50	&	0.579	/	40	&	0.457	/	92	\\
Brox et al. 	&	0.553	/	32	&	0.515	/	71	&	0.546	/	14	&	0.573	/	51	&	0.590	/	22	&	0.526	/	41	&	0.487	/	75	&	0.546	/	56	&	0.641	/	4	\\
DMF\_ROB 	&	0.548	/	33	&	0.545	/	45	&	0.506	/	66	&	0.706	/	11	&	0.518	/	67	&	0.481	/	73	&	0.471	/	81	&	0.580	/	39	&	0.572	/	34	\\
TI-DOFE 	&	0.547	/	34	&	0.422	/	114	&	0.490	/	91	&	0.579	/	48	&	0.528	/	62	&	0.526	/	42	&	0.636	/	25	&	0.587	/	33	&	0.610	/	14	\\
WLIF-Flow 	&	0.542	/	35	&	0.669	/	1	&	0.532	/	33	&	0.552	/	66	&	0.575	/	30	&	0.454	/	93	&	0.571	/	41	&	0.617	/	26	&	0.368	/	131	\\
Ad-TV-NDC 	&	0.542	/	36	&	0.412	/	118	&	0.516	/	56	&	0.746	/	5	&	0.565	/	39	&	0.290	/	138	&	0.518	/	66	&	0.722	/	4	&	0.567	/	36	\\
MLDP\_OF 	&	0.541	/	37	&	0.537	/	50	&	0.495	/	85	&	0.695	/	13	&	0.509	/	73	&	0.559	/	31	&	0.536	/	59	&	0.561	/	47	&	0.434	/	111	\\
TCOF 	&	0.536	/	38	&	0.471	/	93	&	0.506	/	68	&	0.572	/	52	&	0.568	/	35	&	0.578	/	29	&	0.606	/	35	&	0.506	/	71	&	0.483	/	78	\\
Local-TV-L1 	&	0.536	/	39	&	0.530	/	54	&	0.494	/	87	&	0.736	/	8	&	0.653	/	3	&	0.318	/	135	&	0.452	/	88	&	0.494	/	77	&	0.607	/	17	\\
Classic++ 	&	0.535	/	40	&	0.564	/	29	&	0.495	/	86	&	0.713	/	10	&	0.505	/	76	&	0.481	/	71	&	0.466	/	83	&	0.473	/	82	&	0.587	/	28	\\
nLayers 	&	0.529	/	41	&	0.554	/	37	&	0.529	/	41	&	0.602	/	39	&	0.560	/	43	&	0.612	/	22	&	0.444	/	90	&	0.520	/	65	&	0.414	/	119	\\
2DHMM-SAS 	&	0.529	/	42	&	0.574	/	21	&	0.489	/	93	&	0.372	/	106	&	0.565	/	38	&	0.509	/	55	&	0.717	/	14	&	0.550	/	53	&	0.454	/	94	\\
Filter Flow 	&	0.526	/	43	&	0.508	/	75	&	0.453	/	121	&	0.560	/	62	&	0.408	/	114	&	0.676	/	16	&	0.559	/	44	&	0.537	/	60	&	0.504	/	65	\\
MDP-Flow 	&	0.522	/	44	&	0.619	/	5	&	0.518	/	52	&	0.597	/	43	&	0.468	/	94	&	0.522	/	46	&	0.375	/	109	&	0.606	/	31	&	0.473	/	87	\\
JOF 	&	0.521	/	45	&	0.534	/	52	&	0.544	/	17	&	0.646	/	29	&	0.565	/	37	&	0.474	/	78	&	0.322	/	124	&	0.651	/	16	&	0.433	/	112	\\
PH-Flow 	&	0.521	/	46	&	0.450	/	107	&	0.531	/	35	&	0.695	/	14	&	0.549	/	48	&	0.516	/	48	&	0.379	/	108	&	0.624	/	23	&	0.424	/	114	\\
TC-Flow 	&	0.521	/	47	&	0.440	/	108	&	0.472	/	109	&	0.741	/	6	&	0.539	/	59	&	0.448	/	97	&	0.516	/	67	&	0.571	/	43	&	0.441	/	104	\\
NN-field 	&	0.521	/	48	&	0.563	/	32	&	0.506	/	67	&	0.428	/	95	&	0.472	/	89	&	0.543	/	34	&	0.548	/	51	&	0.609	/	30	&	0.496	/	74	\\
Learning Flow 	&	0.521	/	49	&	0.551	/	39	&	0.510	/	62	&	0.495	/	84	&	0.544	/	54	&	0.511	/	52	&	0.540	/	56	&	0.423	/	104	&	0.591	/	24	\\
PGAM+LK 	&	0.519	/	50	&	0.462	/	98	&	0.481	/	100	&	0.274	/	126	&	0.413	/	113	&	0.766	/	4	&	0.626	/	29	&	0.563	/	46	&	0.562	/	39	\\
AGIF+OF 	&	0.518	/	51	&	0.530	/	55	&	0.562	/	8	&	0.578	/	49	&	0.577	/	27	&	0.513	/	51	&	0.422	/	97	&	0.632	/	21	&	0.332	/	140	\\
OFRF 	&	0.517	/	52	&	0.408	/	123	&	0.497	/	82	&	0.478	/	86	&	0.488	/	83	&	0.553	/	33	&	0.754	/	10	&	0.513	/	69	&	0.450	/	99	\\
BlockOverlap 	&	0.517	/	53	&	0.525	/	62	&	0.539	/	21	&	0.544	/	69	&	0.545	/	53	&	0.482	/	70	&	0.351	/	114	&	0.512	/	70	&	0.636	/	6	\\
NNF-EAC 	&	0.513	/	54	&	0.561	/	33	&	0.460	/	117	&	0.460	/	92	&	0.457	/	97	&	0.362	/	130	&	0.769	/	7	&	0.560	/	48	&	0.475	/	85	\\
TriFlow 	&	0.512	/	55	&	0.452	/	105	&	0.498	/	80	&	0.659	/	23	&	0.389	/	125	&	0.458	/	92	&	0.543	/	55	&	0.514	/	68	&	0.588	/	27	\\
ComplOF-FED-GPU 	&	0.509	/	56	&	0.580	/	16	&	0.494	/	88	&	0.388	/	102	&	0.564	/	40	&	0.429	/	113	&	0.466	/	82	&	0.587	/	34	&	0.562	/	40	\\
FlowFields+ 	&	0.505	/	57	&	0.525	/	60	&	0.459	/	118	&	0.562	/	60	&	0.471	/	90	&	0.465	/	85	&	0.533	/	61	&	0.585	/	35	&	0.440	/	105	\\
Sparse-NonSparse 	&	0.505	/	58	&	0.554	/	38	&	0.517	/	53	&	0.649	/	28	&	0.482	/	86	&	0.437	/	109	&	0.427	/	95	&	0.527	/	62	&	0.444	/	101	\\
SILK 	&	0.504	/	59	&	0.427	/	113	&	0.451	/	122	&	0.320	/	115	&	0.545	/	52	&	0.690	/	13	&	0.314	/	127	&	0.689	/	14	&	0.593	/	23	\\
Occlusion-TV-L1 	&	0.503	/	60	&	0.493	/	82	&	0.445	/	127	&	0.740	/	7	&	0.390	/	124	&	0.499	/	62	&	0.462	/	85	&	0.548	/	54	&	0.450	/	97	\\
TF+OM 	&	0.502	/	61	&	0.471	/	94	&	0.558	/	10	&	0.545	/	68	&	0.422	/	110	&	0.288	/	139	&	0.625	/	31	&	0.441	/	95	&	0.667	/	1	\\
OFH 	&	0.500	/	62	&	0.563	/	31	&	0.516	/	55	&	0.504	/	81	&	0.518	/	69	&	0.403	/	119	&	0.555	/	45	&	0.470	/	85	&	0.468	/	88	\\
F-TV-L1 	&	0.500	/	63	&	0.438	/	110	&	0.412	/	137	&	0.757	/	3	&	0.474	/	87	&	0.360	/	131	&	0.544	/	54	&	0.415	/	107	&	0.600	/	22	\\
OFLAF 	&	0.498	/	64	&	0.453	/	103	&	0.552	/	12	&	0.613	/	36	&	0.625	/	10	&	0.526	/	45	&	0.343	/	117	&	0.438	/	97	&	0.437	/	109	\\
TriangleFlow 	&	0.497	/	65	&	0.407	/	125	&	0.519	/	49	&	0.516	/	78	&	0.546	/	51	&	0.707	/	8	&	0.564	/	43	&	0.334	/	125	&	0.385	/	126	\\
3DFlow 	&	0.497	/	66	&	0.457	/	101	&	0.577	/	5	&	0.432	/	94	&	0.539	/	58	&	0.531	/	39	&	0.300	/	132	&	0.584	/	36	&	0.554	/	48	\\
AggregFlow 	&	0.496	/	67	&	0.411	/	121	&	0.535	/	28	&	0.572	/	53	&	0.442	/	102	&	0.529	/	40	&	0.674	/	20	&	0.367	/	119	&	0.439	/	107	\\
S2F-IF 	&	0.496	/	68	&	0.522	/	66	&	0.509	/	63	&	0.569	/	57	&	0.413	/	112	&	0.466	/	84	&	0.432	/	92	&	0.550	/	52	&	0.503	/	66	\\
SRR-TVOF-NL 	&	0.494	/	69	&	0.418	/	115	&	0.536	/	26	&	0.556	/	64	&	0.472	/	88	&	0.464	/	86	&	0.535	/	60	&	0.556	/	49	&	0.415	/	118	\\
COFM 	&	0.493	/	70	&	0.436	/	111	&	0.596	/	1	&	0.594	/	45	&	0.561	/	42	&	0.402	/	120	&	0.309	/	130	&	0.540	/	58	&	0.506	/	63	\\
PMF 	&	0.493	/	71	&	0.511	/	74	&	0.507	/	65	&	0.317	/	118	&	0.563	/	41	&	0.508	/	56	&	0.646	/	23	&	0.538	/	59	&	0.352	/	133	\\
Layers++ 	&	0.493	/	72	&	0.516	/	70	&	0.504	/	70	&	0.651	/	27	&	0.547	/	49	&	0.443	/	101	&	0.236	/	139	&	0.615	/	28	&	0.431	/	113	\\
FESL 	&	0.492	/	73	&	0.483	/	88	&	0.510	/	61	&	0.576	/	50	&	0.541	/	56	&	0.430	/	111	&	0.498	/	71	&	0.498	/	76	&	0.398	/	123	\\
   \hline
\end{tabular}
\end{table*}

\begin{table*}[h]
\caption{Subjective Quality Values and the Re-ranking of the Middlebury Benchmark (Part II). }
\label{tb:value2}
\centering
\begin{tabular}{l |l|l l l l l l l l l  }
  \hline
 & \multicolumn{1}{c}{Average}\vline & \multicolumn{1}{c}{Mequon} & \multicolumn{1}{c}{Schefflera} & \multicolumn{1}{c}{Urban}& \multicolumn{1}{c}{Teddy}& \multicolumn{1}{c}{Backyard}& \multicolumn{1}{c}{Basketball}& \multicolumn{1}{c}{Dumptruck}& \multicolumn{1}{c}{Evergreen}\\
    & value / rank & value / rank &	value / rank	&	value / rank	&	value / rank	&	value / rank	&	value / rank	&	value / rank	&	value / rank\\
        \hline
Classic+NL 	&	0.491	/	74	&	0.585	/	11	&	0.538	/	24	&	0.523	/	77	&	0.558	/	45	&	0.500	/	59	&	0.330	/	121	&	0.482	/	81	&	0.416	/	117	\\
RFlow 	&	0.491	/	75	&	0.466	/	96	&	0.542	/	20	&	0.611	/	37	&	0.449	/	100	&	0.472	/	80	&	0.390	/	105	&	0.439	/	96	&	0.555	/	47	\\
Classic+CPF 	&	0.490	/	76	&	0.528	/	56	&	0.536	/	27	&	0.467	/	90	&	0.592	/	19	&	0.477	/	76	&	0.412	/	101	&	0.543	/	57	&	0.365	/	132	\\
SLK 	&	0.489	/	77	&	0.417	/	117	&	0.427	/	133	&	0.388	/	101	&	0.470	/	92	&	0.640	/	18	&	0.626	/	30	&	0.385	/	117	&	0.563	/	37	\\
FlowNetS+ft+v 	&	0.489	/	78	&	0.524	/	63	&	0.432	/	130	&	0.555	/	65	&	0.573	/	34	&	0.392	/	123	&	0.607	/	34	&	0.276	/	135	&	0.555	/	45	\\
H+S\_ROB 	&	0.488	/	79	&	0.545	/	44	&	0.439	/	128	&	0.307	/	121	&	0.399	/	118	&	0.702	/	9	&	0.455	/	86	&	0.472	/	84	&	0.582	/	29	\\
FOLKI 	&	0.487	/	80	&	0.453	/	104	&	0.490	/	92	&	0.414	/	97	&	0.533	/	61	&	0.311	/	136	&	0.570	/	42	&	0.517	/	67	&	0.610	/	15	\\
DPOF 	&	0.486	/	81	&	0.462	/	99	&	0.524	/	44	&	0.417	/	96	&	0.430	/	108	&	0.402	/	121	&	0.690	/	19	&	0.523	/	63	&	0.443	/	102	\\
HBM-GC 	&	0.486	/	82	&	0.556	/	35	&	0.476	/	103	&	0.665	/	21	&	0.575	/	31	&	0.468	/	81	&	0.245	/	136	&	0.388	/	115	&	0.512	/	62	\\
Fusion 	&	0.486	/	83	&	0.555	/	36	&	0.500	/	75	&	0.571	/	54	&	0.525	/	63	&	0.601	/	24	&	0.342	/	119	&	0.442	/	94	&	0.350	/	136	\\
NL-TV-NCC 	&	0.485	/	84	&	0.546	/	43	&	0.520	/	46	&	0.543	/	70	&	0.389	/	126	&	0.363	/	129	&	0.520	/	64	&	0.457	/	89	&	0.542	/	56	\\
ROF-ND 	&	0.485	/	85	&	0.418	/	116	&	0.485	/	97	&	0.619	/	34	&	0.452	/	99	&	0.500	/	60	&	0.389	/	106	&	0.580	/	38	&	0.437	/	108	\\
Black \& Anandan 	&	0.482	/	86	&	0.523	/	65	&	0.511	/	60	&	0.155	/	140	&	0.591	/	20	&	0.504	/	57	&	0.414	/	100	&	0.553	/	51	&	0.606	/	18	\\
Aniso-Texture 	&	0.481	/	87	&	0.574	/	22	&	0.511	/	59	&	0.584	/	46	&	0.329	/	135	&	0.618	/	21	&	0.281	/	135	&	0.449	/	91	&	0.506	/	64	\\
Sparse Occlusion 	&	0.481	/	88	&	0.527	/	57	&	0.478	/	101	&	0.681	/	19	&	0.392	/	122	&	0.435	/	110	&	0.384	/	107	&	0.448	/	92	&	0.503	/	67	\\
Adaptive 	&	0.480	/	89	&	0.573	/	24	&	0.497	/	81	&	0.651	/	26	&	0.349	/	133	&	0.442	/	103	&	0.315	/	126	&	0.396	/	113	&	0.617	/	13	\\
ACK-Prior 	&	0.477	/	90	&	0.543	/	47	&	0.535	/	29	&	0.269	/	127	&	0.435	/	104	&	0.493	/	67	&	0.519	/	65	&	0.584	/	37	&	0.440	/	106	\\
CPM-Flow 	&	0.477	/	91	&	0.526	/	58	&	0.537	/	25	&	0.382	/	104	&	0.370	/	131	&	0.438	/	107	&	0.578	/	38	&	0.430	/	101	&	0.551	/	51	\\
IAOF2 	&	0.475	/	92	&	0.399	/	129	&	0.464	/	114	&	0.560	/	63	&	0.427	/	109	&	0.393	/	122	&	0.616	/	33	&	0.446	/	93	&	0.496	/	73	\\
Horn \& Schunck 	&	0.474	/	93	&	0.503	/	77	&	0.424	/	135	&	0.264	/	128	&	0.514	/	71	&	0.514	/	50	&	0.538	/	57	&	0.501	/	75	&	0.531	/	58	\\
Correlation Flow 	&	0.473	/	94	&	0.525	/	61	&	0.557	/	11	&	0.527	/	74	&	0.397	/	119	&	0.627	/	20	&	0.329	/	122	&	0.353	/	121	&	0.468	/	89	\\
ProbFlowFields 	&	0.472	/	95	&	0.472	/	92	&	0.512	/	58	&	0.490	/	85	&	0.493	/	81	&	0.440	/	106	&	0.310	/	128	&	0.473	/	83	&	0.590	/	26	\\
ComponentFusion 	&	0.472	/	96	&	0.493	/	81	&	0.521	/	45	&	0.595	/	44	&	0.376	/	130	&	0.696	/	11	&	0.430	/	93	&	0.320	/	127	&	0.341	/	138	\\
SVFilterOh 	&	0.471	/	97	&	0.407	/	124	&	0.546	/	15	&	0.331	/	110	&	0.560	/	44	&	0.480	/	75	&	0.579	/	37	&	0.517	/	66	&	0.351	/	135	\\
Steered-L1 	&	0.471	/	98	&	0.541	/	48	&	0.503	/	73	&	0.294	/	124	&	0.245	/	141	&	0.493	/	66	&	0.554	/	47	&	0.504	/	73	&	0.632	/	8	\\
Ramp 	&	0.470	/	99	&	0.515	/	72	&	0.544	/	16	&	0.470	/	88	&	0.547	/	50	&	0.438	/	108	&	0.343	/	118	&	0.502	/	74	&	0.405	/	121	\\
LSM 	&	0.469	/	100	&	0.517	/	69	&	0.514	/	57	&	0.549	/	67	&	0.484	/	84	&	0.467	/	83	&	0.336	/	120	&	0.438	/	98	&	0.450	/	98	\\
FlowNet2 	&	0.469	/	101	&	0.387	/	132	&	0.475	/	105	&	0.525	/	75	&	0.510	/	72	&	0.442	/	102	&	0.634	/	26	&	0.337	/	124	&	0.443	/	103	\\
CNN-flow-warp+ref 	&	0.468	/	102	&	0.603	/	9	&	0.490	/	90	&	0.412	/	98	&	0.483	/	85	&	0.464	/	88	&	0.365	/	111	&	0.306	/	129	&	0.618	/	12	\\
S2D-Matching 	&	0.467	/	103	&	0.559	/	34	&	0.500	/	74	&	0.471	/	87	&	0.553	/	47	&	0.367	/	128	&	0.422	/	96	&	0.400	/	112	&	0.463	/	91	\\
FlowFields 	&	0.466	/	104	&	0.513	/	73	&	0.489	/	94	&	0.570	/	56	&	0.403	/	117	&	0.389	/	124	&	0.429	/	94	&	0.547	/	55	&	0.387	/	125	\\
HCIC-L 	&	0.464	/	105	&	0.410	/	122	&	0.421	/	136	&	0.435	/	93	&	0.393	/	121	&	0.646	/	17	&	0.575	/	39	&	0.387	/	116	&	0.448	/	100	\\
StereoOF-V1MT 	&	0.464	/	106	&	0.457	/	102	&	0.495	/	84	&	0.309	/	120	&	0.418	/	111	&	0.445	/	99	&	0.486	/	76	&	0.468	/	88	&	0.633	/	7	\\
BriefMatch 	&	0.458	/	107	&	0.518	/	68	&	0.499	/	78	&	0.223	/	135	&	0.506	/	75	&	0.482	/	69	&	0.309	/	129	&	0.649	/	17	&	0.480	/	79	\\
EPMNet 	&	0.458	/	108	&	0.402	/	128	&	0.473	/	107	&	0.569	/	58	&	0.393	/	120	&	0.409	/	117	&	0.552	/	48	&	0.411	/	109	&	0.452	/	96	\\
Adaptive flow 	&	0.458	/	109	&	0.377	/	134	&	0.499	/	79	&	0.505	/	80	&	0.522	/	65	&	0.481	/	72	&	0.326	/	123	&	0.571	/	44	&	0.381	/	128	\\
StereoFlow 	&	0.457	/	110	&	0.369	/	136	&	0.464	/	115	&	0.597	/	42	&	0.351	/	132	&	0.503	/	58	&	0.441	/	91	&	0.438	/	99	&	0.497	/	72	\\
2D-CLG 	&	0.454	/	111	&	0.564	/	28	&	0.446	/	126	&	0.324	/	113	&	0.454	/	98	&	0.420	/	115	&	0.361	/	112	&	0.413	/	108	&	0.646	/	3	\\
Nguyen 	&	0.453	/	112	&	0.464	/	97	&	0.396	/	139	&	0.640	/	32	&	0.490	/	82	&	0.354	/	132	&	0.482	/	78	&	0.305	/	130	&	0.494	/	75	\\
HAST 	&	0.453	/	113	&	0.548	/	42	&	0.531	/	37	&	0.398	/	100	&	0.584	/	26	&	0.345	/	133	&	0.409	/	102	&	0.402	/	111	&	0.406	/	120	\\
EPPM w/o HM 	&	0.452	/	114	&	0.439	/	109	&	0.473	/	108	&	0.316	/	119	&	0.379	/	129	&	0.526	/	44	&	0.662	/	22	&	0.428	/	102	&	0.394	/	124	\\
CostFilter 	&	0.452	/	115	&	0.488	/	85	&	0.543	/	18	&	0.383	/	103	&	0.471	/	91	&	0.441	/	104	&	0.521	/	63	&	0.422	/	105	&	0.347	/	137	\\
UnFlow 	&	0.451	/	116	&	0.458	/	100	&	0.449	/	125	&	0.564	/	59	&	0.497	/	78	&	0.423	/	114	&	0.238	/	138	&	0.505	/	72	&	0.475	/	86	\\
Complementary OF 	&	0.450	/	117	&	0.520	/	67	&	0.524	/	43	&	0.321	/	114	&	0.466	/	95	&	0.373	/	126	&	0.454	/	87	&	0.468	/	87	&	0.475	/	84	\\
IAOF 	&	0.445	/	118	&	0.405	/	126	&	0.454	/	119	&	0.277	/	125	&	0.434	/	105	&	0.460	/	90	&	0.523	/	62	&	0.469	/	86	&	0.540	/	57	\\
FC-2Layers-FF 	&	0.445	/	119	&	0.379	/	133	&	0.519	/	48	&	0.570	/	55	&	0.520	/	66	&	0.448	/	96	&	0.299	/	133	&	0.405	/	110	&	0.418	/	116	\\
TVL1\_ROB 	&	0.441	/	120	&	0.495	/	79	&	0.497	/	83	&	0.540	/	72	&	0.470	/	93	&	0.320	/	134	&	0.369	/	110	&	0.276	/	134	&	0.561	/	42	\\
TV-L1-improved 	&	0.438	/	121	&	0.544	/	46	&	0.454	/	120	&	0.215	/	137	&	0.442	/	101	&	0.526	/	43	&	0.418	/	98	&	0.362	/	120	&	0.546	/	53	\\
Pyramid LK 	&	0.435	/	122	&	0.476	/	89	&	0.478	/	102	&	0.210	/	139	&	0.433	/	107	&	0.596	/	25	&	0.483	/	77	&	0.347	/	122	&	0.454	/	93	\\
SegOF 	&	0.434	/	123	&	0.411	/	119	&	0.428	/	132	&	0.357	/	107	&	0.323	/	137	&	0.555	/	32	&	0.498	/	72	&	0.417	/	106	&	0.479	/	80	\\
EpicFlow 	&	0.433	/	124	&	0.491	/	83	&	0.474	/	106	&	0.379	/	105	&	0.391	/	123	&	0.496	/	64	&	0.538	/	58	&	0.344	/	123	&	0.352	/	134	\\
Shiralkar 	&	0.432	/	125	&	0.499	/	78	&	0.469	/	110	&	0.325	/	111	&	0.382	/	128	&	0.406	/	118	&	0.513	/	69	&	0.297	/	132	&	0.563	/	38	\\
Efficient-NL 	&	0.428	/	126	&	0.490	/	84	&	0.461	/	116	&	0.298	/	123	&	0.567	/	36	&	0.429	/	112	&	0.480	/	79	&	0.314	/	128	&	0.382	/	127	\\
HBpMotionGpu 	&	0.426	/	127	&	0.370	/	135	&	0.429	/	131	&	0.536	/	73	&	0.407	/	115	&	0.385	/	125	&	0.170	/	140	&	0.615	/	27	&	0.500	/	70	\\
GraphCuts 	&	0.426	/	128	&	0.342	/	139	&	0.469	/	111	&	0.238	/	130	&	0.507	/	74	&	0.542	/	36	&	0.356	/	113	&	0.493	/	79	&	0.463	/	90	\\
Dynamic MRF 	&	0.421	/	129	&	0.504	/	76	&	0.500	/	76	&	0.236	/	131	&	0.518	/	70	&	0.461	/	89	&	0.320	/	125	&	0.278	/	133	&	0.548	/	52	\\
PGM-C 	&	0.420	/	130	&	0.473	/	90	&	0.504	/	69	&	0.300	/	122	&	0.329	/	136	&	0.445	/	100	&	0.544	/	53	&	0.435	/	100	&	0.333	/	139	\\
2bit-BM-tele 	&	0.412	/	131	&	0.411	/	120	&	0.538	/	23	&	0.340	/	109	&	0.628	/	9	&	0.285	/	140	&	0.344	/	116	&	0.194	/	141	&	0.555	/	46	\\
SimpleFlow 	&	0.408	/	132	&	0.575	/	20	&	0.508	/	64	&	0.224	/	134	&	0.321	/	138	&	0.441	/	105	&	0.416	/	99	&	0.304	/	131	&	0.479	/	81	\\
IIOF-NLDP 	&	0.407	/	133	&	0.403	/	127	&	0.520	/	47	&	0.406	/	99	&	0.437	/	103	&	0.499	/	61	&	0.303	/	131	&	0.200	/	140	&	0.489	/	77	\\
LocallyOriented 	&	0.406	/	134	&	0.452	/	106	&	0.450	/	123	&	0.319	/	117	&	0.344	/	134	&	0.451	/	95	&	0.350	/	115	&	0.377	/	118	&	0.502	/	69	\\
Rannacher 	&	0.388	/	135	&	0.473	/	91	&	0.482	/	99	&	0.233	/	133	&	0.302	/	139	&	0.560	/	30	&	0.241	/	137	&	0.389	/	114	&	0.421	/	115	\\
FFV1MT 	&	0.385	/	136	&	0.351	/	138	&	0.389	/	141	&	0.260	/	129	&	0.388	/	127	&	0.474	/	79	&	0.488	/	74	&	0.236	/	137	&	0.497	/	71	\\
AdaConv-v1 	&	0.381	/	137	&	0.358	/	137	&	0.449	/	124	&	0.215	/	138	&	0.493	/	80	&	0.246	/	141	&	0.465	/	84	&	0.450	/	90	&	0.368	/	130	\\
Heeger++ 	&	0.372	/	138	&	0.336	/	140	&	0.389	/	140	&	0.319	/	116	&	0.405	/	116	&	0.464	/	87	&	0.399	/	103	&	0.229	/	138	&	0.435	/	110	\\
GroupFlow 	&	0.359	/	139	&	0.399	/	130	&	0.434	/	129	&	0.130	/	141	&	0.287	/	140	&	0.447	/	98	&	0.448	/	89	&	0.324	/	126	&	0.403	/	122	\\
SPSA-learn 	&	0.346	/	140	&	0.398	/	131	&	0.500	/	77	&	0.215	/	136	&	0.465	/	96	&	0.296	/	137	&	0.282	/	134	&	0.243	/	136	&	0.368	/	129	\\
Periodicity 	&	0.295	/	141	&	0.297	/	141	&	0.406	/	138	&	0.233	/	132	&	0.434	/	106	&	0.369	/	127	&	0.168	/	141	&	0.215	/	139	&	0.240	/	141	\\
   \hline
\end{tabular}
\end{table*}

\begin{table*}[h]
\caption{Re-ranking of the Middlebury Benchmark. new: re-ranking given by subjective study. old: ranking in the Middlebury benchmark (Part I).}
\label{tb:rankcom1}
\centering
\begin{tabular}{l |l|l l l l l l l l l}
  \hline
  
   & \multicolumn{1}{c}{Average}\vline & \multicolumn{1}{c}{Mequon} & \multicolumn{1}{c}{Schefflera} & \multicolumn{1}{c}{Urban}& \multicolumn{1}{c}{Teddy}& \multicolumn{1}{c}{Backyard}& \multicolumn{1}{c}{Basketball}& \multicolumn{1}{c}{Dumptruck}& \multicolumn{1}{c}{Evergreen}\\
   & new / old &	new / old &	new / old	&	new / old	&	new / old	&	new / old	&	new / old	&	new / old	&	new / old\\
    
    \hline
SuperSlomo 	&	{\color{blue}1	/	5}	&	3	/	2	&	71	/	34	&	1	/	1	&	2	/	2	&	6	/	1	&	8	/	3	&	1	/	1	&	32	/	3	\\
CtxSyn 	&	{\color{blue}2	/	1}	&	2	/	1	&	3	/	1	&	91	/	63	&	1	/	1	&	1	/	2	&	13	/	1	&	2	/	3	&	61	/	2	\\
DeepFlow2 	&	3	/	9	&	26	/	24	&	39	/	57	&	9	/	8	&	23	/	21	&	26	/	39	&	3	/	6	&	18	/	22	&	9	/	31	\\
SuperFlow 	&	4	/	10	&	13	/	15	&	104	/	70	&	82	/	41	&	46	/	40	&	5	/	9	&	4	/	25	&	10	/	31	&	5	/	7	\\
DeepFlow 	&	{\color{blue}5	/	7}	&	51	/	22	&	13	/	56	&	16	/	7	&	17	/	26	&	38	/	40	&	11	/	10	&	8	/	4	&	21	/	30	\\
ALD-Flow 	&	6	/	21	&	27	/	95	&	2	/	48	&	22	/	6	&	15	/	49	&	116	/	26	&	1	/	12	&	12	/	7	&	82	/	74	\\
PMMST 	&	{\color{blue}7	/	4}	&	7	/	10	&	31	/	15	&	17	/	4	&	77	/	20	&	19	/	6	&	5	/	5	&	32	/	9	&	68	/	9	\\
Aniso. Huber-L1 	&	{\color{blue}8	/	13}	&	18	/	16	&	32	/	99	&	35	/	24	&	12	/	10	&	47	/	46	&	24	/	11	&	11	/	23	&	10	/	15	\\
SIOF 	&	9	/	28	&	49	/	38	&	36	/	98	&	24	/	52	&	25	/	28	&	12	/	4	&	21	/	21	&	45	/	16	&	35	/	48	\\
CBF 	&	{\color{blue}10	/	8}	&	17	/	4	&	40	/	66	&	12	/	9	&	14	/	5	&	27	/	3	&	40	/	13	&	29	/	39	&	30	/	12	\\
Bartels 	&	{\color{red}11	/	77}	&	86	/	115	&	22	/	72	&	18	/	22	&	64	/	77	&	7	/	12	&	73	/	102	&	3	/	17	&	33	/	58	\\
IROF++ 	&	{\color{blue}12	/	17}	&	4	/	31	&	6	/	26	&	71	/	64	&	7	/	3	&	77	/	36	&	6	/	45	&	19	/	18	&	83	/	55	\\
LDOF 	&	13	/	41	&	25	/	32	&	98	/	74	&	89	/	65	&	4	/	35	&	2	/	8	&	17	/	50	&	80	/	79	&	41	/	11	\\
RNLOD-Flow 	&	{\color{red}14	/	71}	&	80	/	37	&	4	/	63	&	33	/	72	&	33	/	15	&	10	/	105	&	52	/	92	&	24	/	47	&	44	/	100	\\
2nd-order prior 	&	15	/	24	&	14	/	11	&	72	/	87	&	38	/	28	&	60	/	27	&	23	/	48	&	70	/	27	&	5	/	24	&	16	/	40	\\
SepConv-v1 	&	16	/	6	&	6	/	3	&	95	/	23	&	83	/	50	&	8	/	30	&	68	/	7	&	15	/	4	&	41	/	2	&	2	/	1	\\
DF-Auto 	&	{\color{blue}17	/	20}	&	95	/	14	&	7	/	69	&	41	/	30	&	55	/	22	&	3	/	15	&	28	/	31	&	78	/	52	&	31	/	18	\\
MDP-Flow2 	&	18	/	3	&	19	/	8	&	89	/	10	&	20	/	13	&	13	/	4	&	28	/	5	&	49	/	24	&	13	/	8	&	95	/	24	\\
CLG-TV 	&	{\color{blue}19	/	15}	&	41	/	13	&	96	/	88	&	47	/	17	&	6	/	16	&	74	/	47	&	46	/	9	&	9	/	20	&	11	/	25	\\
FGIK 	&	20	/	2	&	15	/	5	&	113	/	112	&	112	/	113	&	16	/	124	&	15	/	100	&	12	/	81	&	7	/	131	&	59	/	117	\\
IROF-TV 	&	21	/	14	&	53	/	40	&	50	/	40	&	15	/	20	&	11	/	7	&	54	/	37	&	32	/	49	&	61	/	44	&	49	/	16	\\
LME 	&	22	/	16	&	23	/	17	&	54	/	37	&	40	/	40	&	24	/	23	&	49	/	59	&	36	/	40	&	15	/	6	&	76	/	32	\\
TC/T-Flow 	&	23	/	38	&	59	/	83	&	19	/	61	&	4	/	18	&	79	/	79	&	94	/	53	&	2	/	33	&	103	/	92	&	43	/	60	\\
Modified CLG 	&	24	/	35	&	12	/	7	&	134	/	106	&	25	/	49	&	21	/	46	&	14	/	25	&	104	/	48	&	25	/	14	&	20	/	35	\\
CombBMOF 	&	25	/	19	&	87	/	69	&	9	/	20	&	61	/	37	&	18	/	58	&	65	/	22	&	16	/	32	&	50	/	42	&	54	/	26	\\
CRTflow 	&	26	/	48	&	40	/	47	&	34	/	95	&	79	/	48	&	68	/	14	&	82	/	38	&	18	/	20	&	20	/	89	&	25	/	64	\\
p-harmonic 	&	{\color{blue}27	/	25}	&	10	/	26	&	51	/	92	&	31	/	5	&	32	/	59	&	53	/	41	&	80	/	16	&	22	/	29	&	55	/	44	\\
TV-L1-MCT 	&	28	/	46	&	112	/	77	&	38	/	53	&	76	/	75	&	57	/	19	&	63	/	118	&	27	/	19	&	6	/	62	&	19	/	19	\\
OAR-Flow 	&	{\color{blue}29	/	31}	&	64	/	61	&	112	/	51	&	2	/	12	&	5	/	54	&	91	/	64	&	68	/	23	&	64	/	55	&	50	/	57	\\
FMOF 	&	{\color{blue}30	/	27}	&	8	/	71	&	30	/	13	&	108	/	80	&	29	/	62	&	35	/	16	&	9	/	60	&	42	/	26	&	60	/	54	\\
NNF-Local 	&	31	/	12	&	30	/	12	&	42	/	3	&	30	/	11	&	28	/	67	&	37	/	11	&	50	/	53	&	40	/	19	&	92	/	13	\\
Brox et al. 	&	32	/	26	&	71	/	42	&	14	/	50	&	51	/	32	&	22	/	11	&	41	/	18	&	75	/	80	&	56	/	93	&	4	/	6	\\
DMF\_ROB 	&	33	/	40	&	45	/	66	&	66	/	64	&	11	/	105	&	67	/	64	&	73	/	42	&	81	/	18	&	39	/	27	&	34	/	56	\\
TI-DOFE 	&	{\color{red}34	/	103}	&	114	/	110	&	91	/	137	&	48	/	53	&	62	/	123	&	42	/	31	&	25	/	78	&	33	/	65	&	14	/	106	\\
WLIF-Flow 	&	35	/	22	&	1	/	18	&	33	/	35	&	66	/	59	&	30	/	9	&	93	/	21	&	41	/	99	&	26	/	11	&	131	/	38	\\
Ad-TV-NDC 	&	{\color{blue}36	/	34}	&	118	/	85	&	56	/	127	&	5	/	16	&	39	/	86	&	138	/	29	&	66	/	36	&	4	/	33	&	36	/	8	\\
MLDP\_OF 	&	37	/	57	&	50	/	45	&	85	/	71	&	13	/	14	&	73	/	68	&	31	/	80	&	59	/	91	&	47	/	36	&	111	/	62	\\
TCOF 	&	{\color{red}38	/	70}	&	93	/	59	&	68	/	119	&	52	/	46	&	35	/	31	&	29	/	77	&	35	/	54	&	71	/	118	&	78	/	93	\\
Local-TV-L1 	&	39	/	30	&	54	/	25	&	87	/	101	&	8	/	2	&	3	/	25	&	135	/	60	&	88	/	30	&	77	/	30	&	17	/	10	\\
Classic++ 	&	40	/	53	&	29	/	35	&	86	/	75	&	10	/	23	&	76	/	63	&	71	/	91	&	83	/	88	&	82	/	78	&	28	/	68	\\
nLayers 	&	{\color{blue}41	/	44}	&	37	/	33	&	41	/	11	&	39	/	108	&	43	/	57	&	22	/	120	&	90	/	87	&	65	/	21	&	119	/	45	\\
2DHMM-SAS & {\color{blue}42	/	37}	&	21	/	49	&	93	/	77	&	106	/	69	&	38	/	24	&	55	/	76	&	14	/	43	&	53	/	54	&	94	/	63	\\
Filter Flow 	&	43	/	66	&	75	/	60	&	121	/	115	&	62	/	58	&	114	/	81	&	16	/	17	&	44	/	55	&	60	/	67	&	65	/	46	\\
MDP-Flow 	&	44	/	29	&	5	/	6	&	52	/	14	&	43	/	36	&	94	/	50	&	46	/	63	&	109	/	93	&	31	/	34	&	87	/	33	\\
JOF 	&	45	/	32	&	52	/	44	&	17	/	12	&	29	/	29	&	37	/	32	&	78	/	78	&	124	/	94	&	16	/	15	&	112	/	50	\\
PH-Flow 	&	46	/	23	&	107	/	58	&	35	/	6	&	14	/	19	&	48	/	6	&	48	/	14	&	108	/	109	&	23	/	46	&	114	/	39	\\
TC-Flow 	&	{\color{blue}47	/	49}	&	108	/	101	&	109	/	62	&	6	/	10	&	59	/	60	&	97	/	65	&	67	/	46	&	43	/	43	&	104	/	96	\\
NN-field 	&	{\color{red}48	/	11}	&	32	/	23	&	67	/	4	&	95	/	85	&	89	/	90	&	34	/	13	&	51	/	35	&	30	/	10	&	74	/	20	\\
Learning Flow 	&	{\color{red}49	/	124}	&	39	/	63	&	62	/	108	&	84	/	141	&	54	/	107	&	52	/	130	&	56	/	79	&	104	/	74	&	24	/	114	\\
PGAM+LK 	&	{\color{red}50	/	135}	&	98	/	132	&	100	/	126	&	126	/	132	&	113	/	136	&	4	/	103	&	29	/	106	&	46	/	72	&	39	/	107	\\
AGIF+OF 	&	{\color{blue}51	/	51}	&	55	/	55	&	8	/	31	&	49	/	34	&	27	/	36	&	51	/	87	&	97	/	114	&	21	/	48	&	140	/	101	\\
OFRF 	&	{\color{red}52	/	125}	&	123	/	130	&	82	/	111	&	86	/	81	&	83	/	94	&	33	/	132	&	10	/	126	&	69	/	113	&	99	/	122	\\
BlockOverlap 	&	{\color{blue}53	/	56}	&	62	/	21	&	21	/	96	&	69	/	60	&	53	/	47	&	70	/	51	&	114	/	51	&	70	/	37	&	6	/	5	\\
NNF-EAC 	&	{\color{red}54	/	18}	&	33	/	29	&	117	/	29	&	92	/	51	&	97	/	39	&	130	/	62	&	7	/	8	&	48	/	13	&	85	/	29	\\
TriFlow 	&	{\color{red}55	/	89}	&	105	/	127	&	80	/	91	&	23	/	39	&	125	/	104	&	92	/	84	&	55	/	101	&	68	/	61	&	27	/	73	\\
ComplOF-FED-GPU 	&	56	/	45	&	16	/	86	&	88	/	44	&	102	/	103	&	40	/	41	&	113	/	50	&	82	/	22	&	34	/	66	&	40	/	81	\\
FlowFields+ 	&	57	/	65	&	60	/	62	&	118	/	19	&	60	/	89	&	90	/	92	&	85	/	54	&	61	/	74	&	35	/	51	&	105	/	69	\\
Sparse-NonSparse 	&	{\color{blue}58	/	54}	&	38	/	41	&	53	/	27	&	28	/	62	&	86	/	18	&	109	/	107	&	95	/	86	&	62	/	98	&	101	/	78	\\
SILK 	&	{\color{red}59	/	122}	&	113	/	113	&	122	/	135	&	115	/	137	&	52	/	115	&	13	/	99	&	127	/	110	&	14	/	40	&	23	/	66	\\
Occlusion-TV-L1 	&	60	/	78	&	82	/	64	&	127	/	105	&	7	/	3	&	124	/	108	&	62	/	44	&	85	/	58	&	54	/	80	&	97	/	87	\\
TF+OM 	&	{\color{blue}61	/	64}	&	94	/	104	&	10	/	33	&	68	/	27	&	110	/	91	&	139	/	68	&	31	/	57	&	95	/	77	&	1	/	53	\\
OFH 	&	62	/	90	&	31	/	79	&	55	/	78	&	81	/	84	&	69	/	73	&	119	/	81	&	45	/	41	&	85	/	90	&	88	/	108	\\
F-TV-L1	&	{\color{red}63	/	33}	&	110	/	98	&	137	/	97	&	3	/	25	&	87	/	56	&	131	/	32	&	54	/	7	&	107	/	32	&	22	/	14	\\
OFLAF	&	{\color{blue}64	/	61}	&	103	/	51	&	12	/	8	&	36	/	21	&	10	/	13	&	45	/	117	&	117	/	77	&	97	/	121	&	109	/	102	\\
TriangleFlow 	&	{\color{red}65	/	110}	&	125	/	89	&	49	/	90	&	78	/	82	&	51	/	69	&	8	/	93	&	43	/	85	&	125	/	128	&	126	/	137	\\
3DFlow 	&	66	/	91	&	101	/	91	&	5	/	39	&	94	/	77	&	58	/	72	&	39	/	66	&	132	/	125	&	36	/	91	&	48	/	77	\\
AggregFlow 	&	67	/	74	&	121	/	128	&	28	/	55	&	53	/	55	&	102	/	97	&	40	/	19	&	20	/	26	&	119	/	99	&	107	/	79	\\
S2F-IF 	&	68	/	58	&	66	/	93	&	63	/	18	&	57	/	45	&	112	/	80	&	84	/	95	&	92	/	52	&	52	/	45	&	66	/	88	\\
SRR-TVOF-NL 	&	69	/	87	&	115	/	102	&	26	/	65	&	64	/	83	&	88	/	89	&	86	/	67	&	60	/	104	&	49	/	49	&	118	/	92	\\
COFM 	&	{\color{red}70	/	39}	&	111	/	30	&	1	/	22	&	45	/	26	&	42	/	29	&	120	/	24	&	130	/	123	&	58	/	60	&	63	/	41	\\
PMF 	&	71	/	59	&	74	/	65	&	65	/	43	&	118	/	104	&	41	/	65	&	56	/	27	&	23	/	67	&	59	/	84	&	133	/	110	\\
Layers++ 	&	72	/	43	&	70	/	19	&	70	/	2	&	27	/	42	&	49	/	37	&	101	/	135	&	139	/	128	&	28	/	25	&	113	/	34	\\
FESL 	&	73	/	86	&	88	/	70	&	61	/	32	&	50	/	70	&	56	/	61	&	111	/	121	&	71	/	98	&	76	/	101	&	123	/	97	\\
   \hline
\end{tabular}
\end{table*}

\begin{table*}[h]
\caption{Re-ranking of the Middlebury Benchmark. new: re-ranking given by subjective study. old: ranking in the Middlebury benchmark (Part II).}
\label{tb:rankcom2}
\centering
\begin{tabular}{l |l|l l l l l l l l l}
  \hline
  
   & \multicolumn{1}{c}{Average}\vline & \multicolumn{1}{c}{Mequon} & \multicolumn{1}{c}{Schefflera} & \multicolumn{1}{c}{Urban}& \multicolumn{1}{c}{Teddy}& \multicolumn{1}{c}{Backyard}& \multicolumn{1}{c}{Basketball}& \multicolumn{1}{c}{Dumptruck}& \multicolumn{1}{c}{Evergreen}\\
   & new / old &	new / old &	new / old	&	new / old	&	new / old	&	new / old	&	new / old	&	new / old	&	new / old\\
    
    \hline
Classic+NL 	&	{\color{blue}74	/	75}	&	11	/	50	&	24	/	36	&	77	/	91	&	45	/	17	&	59	/	106	&	121	/	96	&	81	/	105	&	117	/	83	\\
RFlow 	&	75	/	88	&	96	/	46	&	20	/	93	&	37	/	92	&	100	/	78	&	80	/	52	&	105	/	66	&	96	/	88	&	47	/	76	\\
Classic+CPF 	&	{\color{blue}76	/	81}	&	56	/	56	&	27	/	42	&	90	/	71	&	19	/	12	&	76	/	127	&	101	/	127	&	57	/	106	&	132	/	119	\\
SLK 	&	{\color{red}77	/	137}	&	117	/	117	&	133	/	125	&	101	/	135	&	92	/	135	&	18	/	133	&	30	/	83	&	117	/	133	&	37	/	129	\\
FlowNetS+ft+v 	&	78	/	62	&	63	/	39	&	130	/	116	&	65	/	38	&	34	/	44	&	123	/	79	&	34	/	15	&	135	/	95	&	45	/	21	\\
H+S\_ROB 	&	{\color{red}79	/	127}	&	44	/	111	&	128	/	114	&	121	/	138	&	118	/	125	&	9	/	101	&	86	/	82	&	84	/	132	&	29	/	126	\\
FOLKI 	&	{\color{red}80	/	119}	&	104	/	125	&	92	/	138	&	97	/	102	&	61	/	134	&	136	/	74	&	42	/	71	&	67	/	59	&	15	/	49	\\
DPOF 	&	{\color{red}81	/	50}	&	99	/	105	&	44	/	7	&	96	/	96	&	108	/	76	&	121	/	43	&	19	/	70	&	63	/	57	&	102	/	72	\\
HBM-GC 	&	{\color{blue}82	/	80}	&	35	/	43	&	103	/	67	&	21	/	31	&	31	/	43	&	81	/	128	&	136	/	136	&	115	/	82	&	62	/	36	\\
Fusion 	&	83	/	73	&	36	/	34	&	75	/	45	&	54	/	43	&	63	/	70	&	24	/	94	&	119	/	120	&	94	/	83	&	136	/	105	\\
NL-TV-NCC 	&	84	/	107	&	43	/	108	&	46	/	86	&	70	/	73	&	126	/	130	&	129	/	23	&	64	/	115	&	89	/	116	&	56	/	99	\\
ROF-ND 	&	85	/	113	&	116	/	80	&	97	/	81	&	34	/	57	&	99	/	126	&	60	/	102	&	106	/	116	&	38	/	70	&	108	/	132	\\
Black \& Anandan 	&	86	/	69	&	65	/	84	&	60	/	129	&	140	/	126	&	20	/	106	&	57	/	33	&	100	/	42	&	51	/	71	&	18	/	17	\\
Aniso-Texture 	&	{\color{red}87	/	117}	&	22	/	53	&	59	/	107	&	46	/	90	&	135	/	132	&	21	/	126	&	135	/	132	&	91	/	76	&	64	/	115	\\
Sparse Occlusion 	&	88	/	60	&	57	/	73	&	101	/	82	&	19	/	15	&	122	/	55	&	110	/	111	&	107	/	38	&	92	/	87	&	67	/	80	\\
Adaptive 	&	89	/	98	&	24	/	88	&	81	/	117	&	26	/	33	&	133	/	75	&	103	/	122	&	126	/	69	&	113	/	115	&	13	/	59	\\
ACK-Prior 	&	90	/	114	&	47	/	97	&	29	/	49	&	127	/	125	&	104	/	95	&	67	/	129	&	65	/	117	&	37	/	81	&	106	/	109	\\
CPM-Flow 	&	{\color{red}91	/	36}	&	58	/	74	&	25	/	21	&	104	/	56	&	131	/	98	&	107	/	55	&	38	/	39	&	101	/	28	&	51	/	61	\\
IAOF2 	&	92	/	109	&	129	/	112	&	114	/	121	&	63	/	74	&	109	/	87	&	122	/	124	&	33	/	112	&	93	/	73	&	73	/	71	\\
Horn \& Schunck 	&	93	/	85	&	77	/	72	&	135	/	130	&	128	/	118	&	71	/	119	&	50	/	34	&	57	/	29	&	75	/	96	&	58	/	42	\\
Correlation Flow 	&	94	/	112	&	61	/	94	&	11	/	100	&	74	/	35	&	119	/	83	&	20	/	123	&	122	/	135	&	121	/	134	&	89	/	120	\\
ProbFlowFields 	&	{\color{red}95	/	55}	&	92	/	68	&	58	/	16	&	85	/	99	&	81	/	33	&	106	/	49	&	128	/	59	&	83	/	56	&	26	/	28	\\
ComponentFusion 	&	96	/	76	&	81	/	109	&	45	/	30	&	44	/	54	&	130	/	52	&	11	/	90	&	93	/	37	&	127	/	120	&	138	/	116	\\
SVFilterOh 	&	{\color{blue}97	/	95}	&	124	/	87	&	15	/	17	&	110	/	116	&	44	/	66	&	75	/	92	&	37	/	119	&	66	/	53	&	135	/	86	\\
Steered-L1 	&	{\color{blue}98	/	97}	&	48	/	20	&	73	/	47	&	124	/	129	&	141	/	113	&	66	/	98	&	47	/	64	&	73	/	69	&	8	/	89	\\
Ramp 	&	99	/	72	&	72	/	54	&	16	/	28	&	88	/	93	&	50	/	8	&	108	/	104	&	118	/	113	&	74	/	104	&	121	/	82	\\
LSM 	&	{\color{red}100	/	68}	&	69	/	57	&	57	/	38	&	67	/	68	&	84	/	45	&	83	/	108	&	120	/	107	&	98	/	102	&	98	/	90	\\
FlowNet2 	&	{\color{blue}101	/	96}	&	132	/	136	&	105	/	80	&	75	/	88	&	72	/	110	&	102	/	70	&	26	/	56	&	124	/	85	&	103	/	52	\\
CNN-flow-warp+ref 	&	102	/	79	&	9	/	9	&	90	/	76	&	98	/	94	&	85	/	99	&	88	/	83	&	111	/	17	&	129	/	112	&	12	/	43	\\
S2D-Matching 	&	103	/	93	&	34	/	82	&	74	/	68	&	87	/	79	&	47	/	38	&	128	/	88	&	96	/	118	&	112	/	63	&	91	/	67	\\
FlowFields 	&	{\color{red}104	/	42}	&	73	/	67	&	94	/	24	&	56	/	86	&	117	/	74	&	124	/	28	&	94	/	73	&	55	/	35	&	125	/	65	\\
HCIC-L 	&	{\color{red}105	/	138}	&	122	/	139	&	136	/	132	&	93	/	111	&	121	/	133	&	17	/	35	&	39	/	139	&	116	/	117	&	100	/	136	\\
StereoOF-V1MT 	&	106	/	115	&	102	/	120	&	84	/	89	&	120	/	117	&	111	/	129	&	99	/	73	&	76	/	84	&	88	/	125	&	7	/	51	\\
BriefMatch 	&	{\color{blue}107	/	106}	&	68	/	90	&	78	/	54	&	135	/	119	&	75	/	121	&	69	/	45	&	129	/	100	&	17	/	58	&	79	/	70	\\
EPMNet 	&	{\color{blue}108	/	108}	&	128	/	137	&	107	/	73	&	58	/	87	&	120	/	138	&	117	/	75	&	48	/	90	&	109	/	86	&	96	/	94	\\
Adaptive flow 	&	109	/	126	&	134	/	123	&	79	/	136	&	80	/	78	&	65	/	116	&	72	/	137	&	123	/	133	&	44	/	38	&	128	/	104	\\
StereoFlow 	&	110	/	129	&	136	/	141	&	115	/	134	&	42	/	47	&	132	/	71	&	58	/	139	&	91	/	140	&	99	/	50	&	72	/	125	\\
2D-CLG 	&	111	/	83	&	28	/	27	&	126	/	118	&	113	/	101	&	98	/	93	&	115	/	97	&	112	/	14	&	108	/	110	&	3	/	23	\\
Nguyen 	&	112	/	92	&	97	/	92	&	139	/	131	&	32	/	44	&	82	/	103	&	132	/	57	&	78	/	68	&	130	/	123	&	75	/	37	\\
HAST 	&	{\color{red}113	/	63}	&	42	/	28	&	37	/	9	&	100	/	122	&	26	/	34	&	133	/	61	&	102	/	131	&	111	/	97	&	120	/	91	\\
EPPM w/o HM 	&	114	/	104	&	109	/	106	&	108	/	52	&	119	/	124	&	129	/	102	&	44	/	20	&	22	/	121	&	102	/	108	&	124	/	98	\\
CostFilter 	&	115	/	94	&	85	/	114	&	18	/	41	&	103	/	110	&	91	/	112	&	104	/	72	&	63	/	47	&	105	/	109	&	137	/	123	\\
UnFlow 	&	116	/	128	&	100	/	131	&	125	/	103	&	59	/	95	&	78	/	85	&	114	/	136	&	138	/	134	&	72	/	41	&	86	/	138	\\
Complementary OF 	&	117	/	101	&	67	/	116	&	43	/	58	&	114	/	133	&	95	/	109	&	126	/	85	&	87	/	44	&	87	/	94	&	84	/	131	\\
IAOF 	&	118	/	100	&	126	/	119	&	119	/	139	&	125	/	112	&	105	/	101	&	90	/	56	&	62	/	61	&	86	/	75	&	57	/	47	\\
FC-2Layers-FF 	&	{\color{red}119	/	84}	&	133	/	78	&	48	/	5	&	55	/	67	&	66	/	48	&	96	/	109	&	133	/	122	&	110	/	111	&	116	/	85	\\
TVL1\_ROB 	&	120	/	99	&	79	/	103	&	83	/	133	&	72	/	76	&	93	/	105	&	134	/	30	&	110	/	63	&	134	/	114	&	42	/	22	\\
TV-L1-improved 	&	121	/	102	&	46	/	48	&	120	/	113	&	137	/	114	&	101	/	42	&	43	/	89	&	98	/	95	&	120	/	124	&	53	/	75	\\
Pyramid LK 	&	122	/	139	&	89	/	134	&	102	/	140	&	139	/	139	&	107	/	141	&	25	/	140	&	77	/	89	&	122	/	136	&	93	/	141	\\
SegOF 	&	123	/	134	&	119	/	118	&	132	/	83	&	107	/	131	&	137	/	137	&	32	/	131	&	72	/	105	&	106	/	135	&	80	/	124	\\
EpicFlow 	&	{\color{red}124	/	67}	&	83	/	75	&	106	/	46	&	105	/	61	&	123	/	111	&	64	/	71	&	58	/	34	&	123	/	64	&	134	/	121	\\
Shiralkar 	&	{\color{blue}125	/	120}	&	78	/	122	&	110	/	104	&	111	/	107	&	128	/	131	&	118	/	116	&	69	/	76	&	132	/	126	&	38	/	134	\\
Efficient-NL 	&	{\color{blue}126	/	82}	&	84	/	36	&	116	/	60	&	123	/	115	&	36	/	53	&	112	/	110	&	79	/	111	&	128	/	107	&	127	/	111	\\
HBpMotionGpu 	&	127	/	116	&	135	/	124	&	131	/	128	&	73	/	66	&	115	/	114	&	125	/	113	&	140	/	108	&	27	/	12	&	70	/	84	\\
GraphCuts 	&	128	/	105	&	139	/	126	&	111	/	59	&	130	/	127	&	74	/	88	&	36	/	96	&	113	/	72	&	79	/	100	&	90	/	113	\\
Dynamic MRF 	&	129	/	118	&	76	/	81	&	76	/	79	&	131	/	106	&	70	/	118	&	89	/	58	&	125	/	124	&	133	/	119	&	52	/	112	\\
PGM-C 	&	{\color{red}130	/	52}	&	90	/	76	&	69	/	25	&	122	/	98	&	136	/	100	&	100	/	69	&	53	/	28	&	100	/	68	&	139	/	103	\\
2bit-BM-tele 	&	{\color{blue}131	/	130}	&	120	/	99	&	23	/	110	&	109	/	120	&	9	/	84	&	140	/	82	&	116	/	137	&	141	/	141	&	46	/	27	\\
SimpleFlow 	&	132	/	121	&	20	/	52	&	64	/	84	&	134	/	136	&	138	/	51	&	105	/	119	&	99	/	129	&	131	/	137	&	81	/	133	\\
IIOF-NLDP 	&	{\color{blue}133	/	131}	&	127	/	107	&	47	/	85	&	99	/	100	&	103	/	96	&	61	/	125	&	131	/	130	&	140	/	140	&	77	/	128	\\
LocallyOriented 	&	134	/	111	&	106	/	96	&	123	/	122	&	117	/	109	&	134	/	117	&	95	/	114	&	115	/	75	&	118	/	103	&	69	/	95	\\
Rannacher 	&	135	/	123	&	91	/	100	&	99	/	120	&	133	/	121	&	139	/	82	&	30	/	112	&	137	/	97	&	114	/	122	&	115	/	118	\\
FFV1MT 	&	{\color{blue}136	/	133}	&	138	/	133	&	141	/	124	&	129	/	123	&	127	/	127	&	79	/	86	&	74	/	103	&	137	/	127	&	71	/	127	\\
AdaConv-v1 	&	{\color{red}137	/	47}	&	137	/	121	&	124	/	94	&	138	/	97	&	80	/	122	&	141	/	10	&	84	/	2	&	90	/	5	&	130	/	4	\\
Heeger++ 	&	{\color{blue}138	/	136}	&	140	/	135	&	140	/	123	&	116	/	128	&	116	/	128	&	87	/	134	&	103	/	65	&	138	/	129	&	110	/	130	\\
GroupFlow 	&	{\color{blue}139	/	140}	&	130	/	138	&	129	/	109	&	141	/	130	&	140	/	139	&	98	/	138	&	89	/	138	&	126	/	130	&	122	/	135	\\
SPSA-learn 	&	140	/	132	&	131	/	129	&	77	/	102	&	136	/	134	&	96	/	120	&	137	/	115	&	134	/	62	&	136	/	138	&	129	/	140	\\
Periodicity 	&	{\color{blue}141	/	141}	&	141	/	140	&	138	/	141	&	132	/	140	&	106	/	140	&	127	/	141	&	141	/	141	&	139	/	139	&	141	/	139	\\
   \hline
\end{tabular}
\end{table*}

\end{document}